%% file: main_arxiv.tex
\documentclass[10pt,twocolumn,letterpaper]{article}

\usepackage{cvpr}
\usepackage{times}
\usepackage{epsfig}
\usepackage{graphicx}
\usepackage{amsmath}
\usepackage{amssymb}

\input{math_commands.tex}

\usepackage{amsfonts}
\usepackage{array}
\usepackage{booktabs}
\usepackage{mathtools}
\usepackage{microtype}
\usepackage{multirow}
\usepackage{nicefrac}
\usepackage{url}
\usepackage{amsfonts}
\usepackage{xcolor}
\usepackage{pifont}
\usepackage{float}
\usepackage{tabulary}

\usepackage{graphbox}

\newcommand{\cmark}{\ding{52}}%
\newcommand{\xmark}{\ding{56}}%

\newcommand{\vect}[1]{\mathbf{#1}}

\def\eg{\textit{e.g.}~}
\def\ie{\textit{i.e.}}

\newcolumntype{x}[1]{>{\centering\arraybackslash}p{#1pt}}
\newcolumntype{Y}{>{\centering\arraybackslash}X}

\usepackage{tabularx}
\newcolumntype{Y}{>{\centering\arraybackslash}X}

\usepackage{graphicx,overpic,xcolor}
\usepackage{longtable}
\usepackage{colortbl}
\usepackage[caption=false]{subfig}
\usepackage{wrapfig}
\usepackage{caption}

\usepackage{wrapfig}

\newlength\savewidth\newcommand\shline{\noalign{\global\savewidth\arrayrulewidth
  \global\arrayrulewidth 1pt}\hline\noalign{\global\arrayrulewidth\savewidth}}
  
\newcommand{\tablestyle}[2]{\setlength{\tabcolsep}{#1}\renewcommand{\arraystretch}{#2}\centering\footnotesize}

\makeatletter\renewcommand\paragraph{\@startsection{paragraph}{4}{\z@}
  {.5em \@plus1ex \@minus.2ex}{-.5em}{\normalfont\normalsize\bfseries}}\makeatother

\usepackage[pagebackref=true,breaklinks=true,letterpaper=true,colorlinks,bookmarks=false]{hyperref}

\cvprfinalcopy 

\ifcvprfinal\fi
\begin{document}

\title{Dynamic Graph Message Passing Networks}

\author{Li Zhang\textsuperscript{1}
	\and
	Dan Xu\textsuperscript{1}
	\and
	Anurag Arnab\textsuperscript{2}\thanks{Work primarily done at the University of Oxford.}
	\and
	Philip H.S. Torr\textsuperscript{1}
	\and
	\textsuperscript{1}University of Oxford
	\quad
	\textsuperscript{2}Google Research\\
	{\tt\small \{lz, danxu, phst\}@robots.ox.ac.uk \quad aarnab@google.com}
}

\maketitle
\thispagestyle{empty}

\input{1-abstract.tex}
\input{2-introduction.tex}
\input{3-related.tex}

\input{4-method.tex}
\input{5-experiments.tex}
\input{6-conclusion.tex}
\input{acknowledge}

\clearpage
\appendix

\section*{Appendix}
\input{7-supp.tex}

\clearpage
{\small
\bibliographystyle{ieee_fullname}
\bibliography{biblio}
}

\end{document}

%% file: math_commands.tex
\usepackage{amsmath,amsfonts,bm}

\def\eqref#1{equation~\ref{#1}}

\def\1{\bm{1}}

\newcommand{\test}{\mathcal{D_{\mathrm{test}}}}

\DeclareMathAlphabet{\mathsfit}{\encodingdefault}{\sfdefault}{m}{sl}
\SetMathAlphabet{\mathsfit}{bold}{\encodingdefault}{\sfdefault}{bx}{n}



%% file: 1-abstract.tex
\begin{abstract}
Modelling long-range dependencies is critical for scene understanding tasks in computer vision.
Although CNNs have excelled in many vision tasks, they are still limited in capturing long-range structured relationships as they typically consist of layers of local kernels.
A fully-connected graph is beneficial for such modelling, however, its computational overhead is prohibitive.
We propose a dynamic graph message passing network, 
that significantly reduces the computational complexity compared to related works modelling a fully-connected graph.
This is achieved by adaptively sampling nodes in the graph, conditioned on the input, for message passing.
Based on the sampled nodes, we dynamically predict node-dependent filter weights and the affinity matrix for propagating information between them.
Using this model, 
we show significant
improvements with respect to strong, state-of-the-art baselines on three different
tasks and backbone architectures. Our approach also outperforms fully-connected
graphs while using substantially fewer floating-point operations and parameters.
The project website is~\url{http://www.robots.ox.ac.uk/~lz/dgmn/}.
\end{abstract}

%% file: 2-introduction.tex
\section{Introduction}
\label{sec:intro}
Capturing long-range dependencies is crucial for complex scene understanding tasks such as semantic segmentation, instance segmentation and object detection.
Although convolutional neural networks (CNNs) have excelled in a wide range of scene understanding tasks \cite{alexnet, vgg, resnet}, they are still limited by their ability to capture these long-range interactions. 
To improve the capability of CNNs in this regard, a recent, popular model Non-local networks~\cite{wang2018nonlocal} proposes a generalisation of the attention model of~\cite{vaswani2017attention} and achieves significant advance in several computer vision tasks. 

Non-local networks essentially model pairwise structured relationships among all feature elements in a feature map to produce the attention weights which are used for feature aggregation.
Considering each feature element as a node in a graph, Non-local networks effectively model a fully-connected feature graph and thus have a quadratic inference complexity with respect to the number of the feature elements.
This is infeasible for dense prediction tasks on high-resolution imagery, as commonly encountered in semantic segmentation~\cite{Cityscapes}.
Moreover, in dense prediction tasks, capturing relations between all pairs of pixels is usually unnecessary due to the redundant information contained within the image (Fig.~\ref{fig:teaser}).
Simply subsampling the feature map to reduce the memory requirements is also suboptimal, as such na\"ive subsampling would result in smaller objects in the image not being represented adequately.

Graph convolution networks (GCNs)~\cite{kipf2016semi, gilmer2017neural} -- which propagate information along graph-structured input data -- can alleviate the computational issues of non-local networks to a certain extent.
However, this stands only if local neighbourhoods are considered for each node.
Employing such local-connected graphs means that the long-range contextual information needed for complex vision tasks such as segmentation and detection~\cite{rabinovich2007objects,oliva2007, bai2009geodesic} 
will only be partially captured.
Along this direction, GraphSAGE~\cite{hamilton2017inductive} introduced an efficient graph learning model based on graph sampling.
However, the proposed sampling method considered a uniform sampling strategy along the spatial dimension of the input, and was independent of the actual input.
Consequently, the modelling capacity was restricted as it assumed a static input graph where the neighbours for each node were fixed and filter weights were shared among all nodes.

\input{figures/teaser}

To address the aforementioned shortcomings, we propose a novel dynamic graph message passing network (DGMN) model, targeting effective and efficient deep representational learning with joint modeling of two key dynamic properties as illustrated in Fig.~\ref{fig:teaser}. Our contribution is twofold: (i) We dynamically sample the neighbourhood of a node from the feature graph, conditioned on the node features. 
Intuitively, this learned sampling allows the network to efficiently gather long-range context by only selecting a subset of the most relevant nodes in the graph; 
(ii) Based on the nodes that have been sampled, we further dynamically predict node-dependant, and thus \textit{position specific}, filter weights and also the affinity matrix, which are used to propagate information among the feature nodes via message passing. 
The dynamic weights and affinities are especially beneficial to specifically model each sampled feature context, leading to more effective message passing.
Both of these dynamic properties are jointly optimised in a single model, and we modularise the DGMN as a network layer for simple deployment into existing networks.

We demonstrate the proposed model on the tasks of semantic segmentation, object detection and instance segmentation on the challenging Cityscapes~\cite{Cityscapes} and COCO~\cite{lin2014microsoft} datasets.
We achieve significant performance improvements over the fully-connected Non-local model~\cite{wang2018nonlocal}, while using substantially fewer floating point operations (FLOPs).
Significantly, one variant of our model with dynamic filters and affinities (\ie,~the second dynamic property) achieves similar performance to Non-local while only using 9.4\% of its FLOPs and 25.3\% of its parameters.
Furthermore, ``plugging'' our module into existing networks, we show considerable improvements with respect to strong, state-of-the-art baselines on three different tasks and backbone architectures.

%% file: figures/teaser.tex
\begin{figure*}
	\centering
	\includegraphics[width=0.92\linewidth]{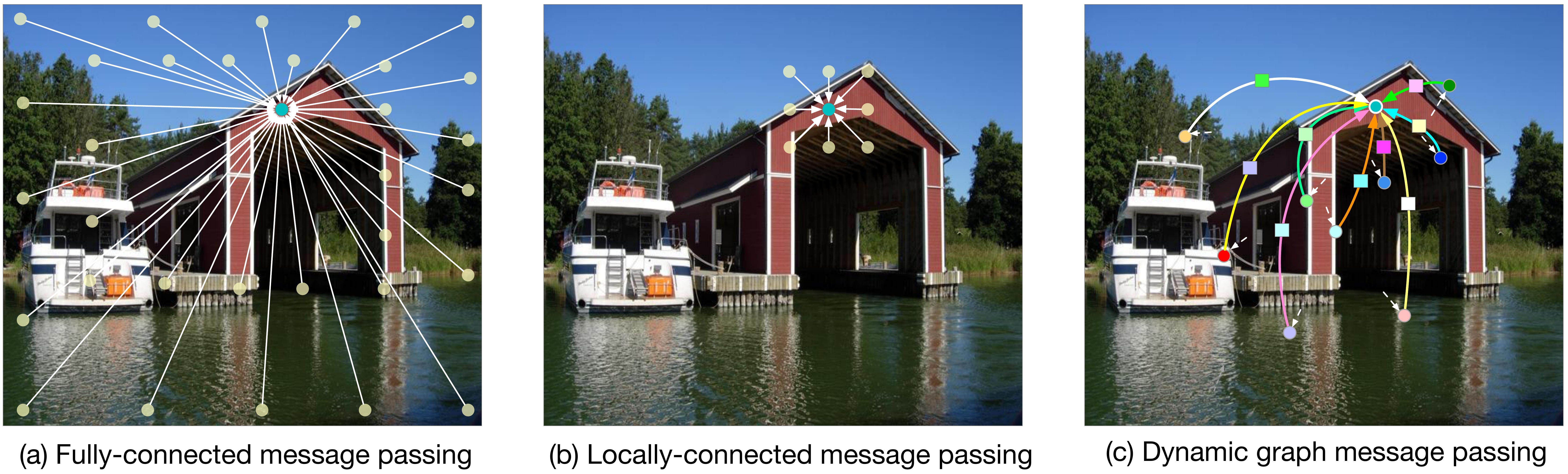}
	\vspace{-0.4\baselineskip}
	\caption{\small Contextual information is crucial for complex scene understanding tasks. To recognise the ``boathouse'', one needs to consider the ``boat'' and the ``water'' next to it.
	Fully-connected message passing models (a) are able to obtain this information, but are prohibitively expensive. Furthermore, they capture a lot of redundant information (\ie ``trees'' and ``sky'').
	Locally-connected models (b) are more efficient, but miss out on important context.
	Our proposed approach (c), dynamically samples a small subset of relevant feature nodes based on a \textit{learned} dynamic sampling scheme, 
	\ie~the \textit{learned} position-specific random walk (indicated by the white dashed arrow lines), and also dynamically predicts filter weights and affinities (indicated by unique edge and square colors.), which are both conditioned on the sampled feature nodes.
	}
	\label{fig:teaser}
 	\vspace{-1\baselineskip}
\end{figure*}

%% file: 3-related.tex
\section{Related work}
\label{sec:related}

An early technique for modelling context for computer vision tasks involved conditional random fields.
In particular, the DenseCRF model~\cite{krahenbuhl2011efficient} was popular as it modelled interactions between all pairs of pixels in an image.
Although such models have been integrated into neural networks~\cite{zheng2015conditional,arnab2016higher,arnab2018conditional, xu2017learning}, they are limited by the fact that the pairwise potentials are based on simple handcrafted features,
Moreover, they mostly model discrete label spaces, and are thus not directly applicable in the feature learning task since feature variables are typically continuous. 
Coupled with the fact that CRFs are computationally expensive, CRFs are no longer used for most computer vision tasks.

A complementary technique for increasing the receptive field of CNNs was to use dilated convolutions~\cite{deeplabv1,yu2015multi}.
With dilated convolutions, the number of parameters does not change, while the receptive field grows exponentially if the dilation rate is linearly increased in successive layers.
Other modifications to the convolution operation include deformable convolution~\cite{dai2017deformable,zhu2018deformable}, which learns the offset with respect to a predefined grid from which to select input values.
However, the weights of the deformable
convolution filters do not depend on the selected input, and are in fact shared across all different positions.
In contrast, our dynamic sampling aims to sample over the whole feature graph to obtain a large receptive field, and the predicted affinities and the weights for message passing are \textit{position specific} and \textit{conditioned} on the dynamically sampled nodes. 
Our model is thus able to better capture position-based semantic context to enable more effective message passing among feature nodes. 

The idea of sampling graph nodes has previously been explored in GraphSAGE~\cite{hamilton2017inductive}.
Crucially, 
GraphSAGE simply uniformly samples nodes. 
In contrast, our sampling strategy is \textit{learned} based on the node features. 
Specifically, we first sample the nodes uniformly in the spatial dimension, and then dynamically predict \textit{walks} of each node conditioned on the node features. 
Furthermore, GraphSAGE does not consider our second important property, \ie,~the dynamic prediction of the affinities and the message passing kernels.

We also note that \cite{jia2016dynamic} developed an idea of ``dynamic convolution'', that is predicting a dynamic convolutional filter for each feature position.
More recently,~\cite{wu2019pay} further reduced the complexity of this operation in the context of natural language processing with lightweight grouped convolutions.
Unlike~\cite{jia2016dynamic,wu2019pay}, we present a graph-based formulation, and jointly learn dynamic weights and dynamic affinities, which are conditioned on an \emph{adaptively sampled} neighbourhood for each feature node in the graph using the proposed dynamic sampling strategy for effective message passing.

%% file: 4-method.tex
\section{Dynamic graph message passing networks}
\label{sec:method}
\subsection{Problem definition and notation}

Given an input feature map interpreted as a set of feature vectors, \ie,~$\vect{F} = \{\vect{f}_i\}_{i=1}^N$  with $\vect{f}_i \in \mathbb{R}^{1\times C}$, where $N$ is the number of pixels and $C$ is the feature dimension, our goal is to learn a set of refined latent feature vectors $\vect{H} = \{\vect{h}_i\}_{i=1}^N$ by utilising hidden structured information among the feature vectors at different pixel locations.
$\vect{H}$ has the same dimension as the observation $\vect{F}$. To learn such structured representations, we convert the feature map into a graph domain by constructing a feature graph $\mathcal{G}=\{\mathcal{V}, \mathcal{E}, A\}$ with $\mathcal{V}$ as its nodes, $\mathcal{E}$ as its edges and $A$ as its adjacency matrix.
Specifically, the nodes of the graph are represented by the latent feature vectors, \ie,~$\mathcal{V} = \{\vect{h}_i\}_{i=1}^N$, and $A \in \mathbb{R}^{N\times N}$ is a binary or learnable matrix with self-loops describing the connections between nodes.
In this work, we propose a novel dynamic graph message passing network~\cite{gilmer2017neural} for deep representation learning, which refines each graph feature node by passing messages on the graph $\mathcal{G}$.
Different from existing message passing neural networks considering a fully- or locally-connected static graph~\cite{wang2018nonlocal, gilmer2017neural}, we propose a dynamic graph network model with two dynamic properties, \ie,~\textit{dynamic sampling} of graph nodes to approximate the full graph distribution, 
and \textit{dynamic prediction} of node-conditioned filter weights and affinities, in order to achieve more efficient and effective message passing.

\subsection{Graph message passing neural networks for deep representation learning}
\label{sec:method_mpgraph}

Message passing neural networks (MPNNs)~\cite{gilmer2017neural} present a generalised form of graph neural networks such as graph convolution networks~\cite{kipf2016semi}, gated graph sequential networks~\cite{li2015gated} 
and graph attention networks~\cite{velivckovic2017graph}.
In order to model structured graph data, in which latent variables are represented as nodes on an undirected or directed graph, feed-forward inference is performed through a message passing phase followed by a readout phase upon the graph nodes.
The message passing phase usually takes $T$ iteration steps to update feature nodes, while the readout phase is for the final prediction, \eg, graph classification with updated nodes.
In this work, we focus on the message passing phase for learning efficient and effective feature refinement, since well-represented features are critical in all downstream tasks.
The message passing phase consists of two steps,  \ie,~a message calculation step $M^t$ and a message updating step $U^t$.
Given a latent feature node $\vect{h}_i^{(t)}$ at an iteration $t$, for computational efficiency, we consider a locally connected node field with $v_i \subset \mathcal{V}$ and $v_i \in \mathbb{R}^{(K\times C)}$, where $K \ll N$ is the number of sampled nodes in $v_i$. Thus we can define the message calculation step for node $i$ operated locally as
\begin{align}
\vect{m}^{(t+1)}_{i} &= M^t \left(A_{i,j},  \{\vect{h}_1^{(t)}, \cdots,  \vect{h}_K^{(t)} \}, \vect{w}_j \right) \nonumber \\ 
&= \sum_{j \in \mathcal{N}(i)} A_{i,j} \vect{h}_j^{(t)} \vect{w}_j, \ \,
\label{equ:messagecalculation}
\end{align}
where $A_{i,j} = A[i, j]$ describes the connection relationship \ie,~the affinity between latent nodes $\vect{h}_i^{(t)}$ and $\vect{h}_j^{(t)}$, $\mathcal{N}(i)$ denotes a self-included neighborhood of the node $\vect{h}_i^{(t)}$ which can be derived from $v_i$ and $\vect{w}_j \in \mathbb{R}^{C \times C}$ is a transformation matrix for message calculation on the hidden node $\vect{h}_j^{(t)}$.
The message updating function $U^t$ then updates the node $\vect{h}_i^{(t)}$ with a linear combination of the calculated message and the observed feature $\vect{f}_i$ at the node position $i$ as: 
\begin{equation}
\vect{h}_i^{(t+1)} = U^t \left(\vect{f}_i, \vect{m}^{(t+1)}_{i} \right) = \sigma \left( \vect{f}_i+ \alpha^m_i \vect{m}^{(t+1)}_{i} \right),
\label{messageupdating}
\end{equation}
where $\alpha^m_i$ of a learnable parameter for scaling the message, and the operation $\sigma(\cdot)$ is a non-linearity function, \eg, ReLU. By iteratively performing message passing on each node with $T$ steps, we obtain a refined feature map $\vect{H}^{(T)}$ as output.

\input{figures/framework}

\subsection{From a fully-connected graph to a dynamic sampled graph}
\label{sec:method_sampling}

A fully-connected graph typically contains many connections and parameters, which, in addition to computational overhead, results in redundancy in the connections, and also makes the network optimisation more difficult especially when dealing with limited training data.
Therefore, as in Eq.~\ref{equ:messagecalculation}, a local node connection field is considered in the graph message passing network. 
However, in various computer vision tasks, such as detection and segmentation, learning deep representations capturing both local and global receptive fields is important for the model performance~\cite{rabinovich2007objects,oliva2007,li2019global,hou2020strip}.
To maintain a large receptive field while utilising much fewer parameters than the fully-connected setting, we further explore dynamic sampling strategies in our proposed graph message passing network.
We develop a uniform sampling scheme, which we then extend to a predicted random walk sampling scheme, aimed at reducing the redundancy found in a fully-connected graph.
This sampling is performed in a dynamic fashion, meaning that for a given node $\vect{h}_i$, we aim to sample an optimal subset of $v_i$ from $\mathcal{V}$ to update $\vect{h}_i$ via message passing as shown in Fig.~\ref{fig:framework}.

\noindent\textbf{Multiple uniform sampling for dynamic receptive fields.} 
Uniform sampling is a commonly used strategy for graph node sampling~\cite{leskovec2006sampling} based on Monte-Carlo estimation. %
To approximate the distribution of $\mathcal{V}$, we consider a set of $S$ uniform sampling rates $\vect{\varphi}$ with $\vect{\varphi} = \{\rho_q\}_{q=1}^S$, where $\rho_q$ is a sampling rate.
Let us assume that the latent feature nodes are located in a $P$-dimensional space $\mathbb{R}^{P}$. 
For instance, $P = 2$  for images considering the $x$- and $y$-axes. 
For each latent node $\vect{h}_i$, a total of $K$ neighbouring nodes are sampled from $\mathbb{R}^P$. 
The receptive field of $v_i$ is thus determined by $\rho_q$ and $K$.
Note that the sampling rate $\rho_q$ corresponds to the ``dilation rate'' often used in convolution~\cite{yu2015multi} and is thus able to capture a large receptive field whilst maintaining a small number of connected nodes.
Thus we can achieve much lower computational overhead compared with fully-connected message passing in which typically all $N$ nodes are used when one of the nodes is updated. 
Each node receives $S$ complementary messages from distinct receptive fields for updating as
\begin{align}
\vect{m}^{(t+1)}_{i} &= \sum_{q} \sum_{j \in \mathcal{N}_{q}(i)} \beta_q A_{i,j}^{q} \vect{h}_j^{(t)} \vect{w}_j^q, 
\label{equ:uniformsampling}
\end{align}
where $\beta_q$ is a weighting parameter for the message from the $q$-th sampling rate and $q = 1, \cdots, S$.
$A^q$ denotes an adjacency matrix formed under a sampling rate $\rho_q$, with 
$A_{i,j}^{q}$, $\vect{w}_j^q$ and $\mathcal{N}_{q}(i)$ defined analogously. 
The uniform sampling scheme acts as a linear sampler based on the spatial distribution while not considering the original feature distribution of the hidden nodes, \ie,~sampling independently of the node features. 
Eq.~\ref{messageupdating} can still be used to update the nodes.

\noindent\textbf{Learning position-specific random walks for node-dependant adaptive sampling.} 
To take into account the feature data distribution when sampling nodes, we further present a random walk strategy upon the uniform sampling. 
Walks around the uniformly sampled nodes could sample the graph in a non-linear and adaptive manner, and we believe that it could facilitate learning better approximation of the original feature distribution. 
The ``random'' here refers to the fact that the walks are predicted in a data-driven fashion from stochastic gradient descent.
Given a matrix, $v_i^q \in \mathbb{R}^{K\times C}$, constructed from $K$ uniformly sampled nodes under a sampling rate, $\rho_q$, the random walk of each node is further estimated based on the feature data of the sampled nodes.
Given the $P$-dimensional space where the nodes distribute ($P=2$ for images), let us denote $\bigtriangleup \vect{d}_j^q \in \mathbb{R}^{P\times 1}$ as predicted walks from a uniformly sampled node $\vect{h}_j$ with $j \in \mathcal{N}_{q}(i)$. 
The node walk prediction can then be performed using a matrix transformation as
\begin{equation}
\bigtriangleup \vect{d}_j^q = \vect{W}_{i,j}^q v_i^q + \vect{b}_{i,j}^q,
\label{equ:deformablewalk}
\end{equation}
where $\vect{W}_{i,j}^q \in \mathbb{R}^{P\times (K\times C)}$ and $\vect{b}_{i,j}^q \in \mathbb{R}^{P\times 1}$ are the matrix transformation parameters, which are learned separately for each node $v_i^q$. 
With the predicted walks, we can obtain a new set of adaptively sampled nodes ${v'}_i^{q}$, 
and generate the corresponding adjacency matrix ${A'}^q$, which can be used to calculate the messages as
\begin{align}
\vect{m}^{(t+1)}_{i} &= \sum_{q} \sum_{j \in \mathcal{N}_{q}(i)} \beta_q {A'}_{i,j}^{q} \varrho \left(\vect{h'}_{j}^{(t)} | \mathcal{V}, j, \bigtriangleup \vect{d}_j^q \right) \vect{w}_j^q, 
\label{equ:deformablemessage}
\end{align}
where the function $\varrho(\cdot)$ is a bilinear sampler~\cite{jaderberg2015spatial} which samples a new feature node $\vect{h'}_{j}^{(t)}$ around $\vect{h}_{j}^{(t)}$ given the predicted walk $\bigtriangleup \vect{d}_j^q$ and the whole set of graph vertexes $\mathcal{V}$.

\input{figures/dmc}

\subsection{Joint learning of node-conditioned dynamic filters and affinities}
\label{sec:method_dynamic_weights}

In the message calculation formulated in Eq.~\ref{equ:deformablemessage}, the set of weights $\{\vect{w}_j^q\}_{j=1}^K$ of the filter is shared for each adaptively sampled node field ${v'}_i^q$. 
However, since each ${v'}_i^q$ essentially defines a node-specific local feature context, it is more meaningful to use a node-conditioned filter to learn the message for each hidden node $\vect{h'}_i^{(t)}$. 
In additional to the filters for the message calculation in Eq.~\ref{equ:deformablemessage}, the affinity ${A'}_{i,j}$ of any pair of nodes $\vect{h'}_i^{(t)}$ and $\vect{h'}_j^{(t)}$ could be also be predicted and should also be conditioned on the node field ${v'}_i^q$, since the affinity reweights the message passing only  in ${v'}_i^q$. 
As shown in Fig.~\ref{fig:dmc}, we thus use matrix transformations to simultaneously estimate the dynamic filter and affinity which are both conditioned on ${v'}_i^q$,
\begin{equation}
\{\vect{w}_j^q, {A'}_{i,j}^q\} = \vect{W}_{i,j}^{k,A} {v'}_i^q + \vect{b}_{i,j}^{k,A}, 
\label{equ:dynamicweights}
\end{equation} 
\begin{equation}
{A'}_{i,j}^q = \mathrm{softmax}_{c} ({A'}_{i,j}^q) =\dfrac{\exp({{A'}_{i,j}^q})}{\sum_{l \in \mathcal{N}_q(i)} \exp({{A'}_{i,l}^q})},
\label{equ:aff_softmax}
\end{equation} 
where the function $\mathrm{softmax}_c(\cdot)$ denotes a softmax operation along the channel axis, which is used to perform a normalisation on the estimated affinity ${A'}_{i,j}^q \in \mathbb{R}^1$. 
$\vect{W}_{i,j}^{k,A} \in \mathbb{R}^{(G\times C + 1)\times (K \times C)}$ and $\vect{b}_{i,j}^{k,A} \in \mathbb{R}^{(G\times C + 1)}$ are matrix transformation parameters.
To reduce the number of the filter parameters, we consider grouped convolutions~\cite{chollet2017xception} with a set of $G$ groups split from the total $C$ feature channels, and $G \ll C$, \ie,~each group of $C/G$ feature channels shares the same set of filter parameters. 
The predicted dynamic filter weights and the affinities are then used in Eq.~\ref{equ:deformablemessage} for dynamic message calculation.

\subsection{Modular instantiation}
Figures~\ref{fig:framework} and \ref{fig:dmc} shows how our proposed dynamic graph message passing network (DGMN) can be implemented in a neural network.
The proposed module accepts a single feature map $\vect{F}$ as input, which can be derived from any CNN layer.
$\vect{H}^{(0)}$ denotes an initial state of the latent feature map, $\vect{H}$, and is initialised with $\vect{F}$. 
$\vect{H}$ and $\vect{F}$ have the same dimension, \ie,~$\vect{F}, \vect{H} \in \mathbb{R}^{H\times W \times C}$, where $H$, $W$ and $C$ are the height, width and the number of feature channels of the feature map respectively.
We first define a set of $S$ uniform sampling rates (we show two uniform sampling rates in Fig.~\ref{fig:framework} for clarity).
The uniform and the random walk sampler sample the nodes from the full graph and return the node indices for subsequent dynamic message calculation (DMC) in Fig.~\ref{fig:dmc}. 
The matrix transformation $\vect{W}_{i,j}^q$ to estimate the node random walk in Eq.~\ref{equ:deformablewalk} is implemented by a $3 \times 3$ convolution layer~\cite{dai2017deformable}.
Note that other sampling strategies could also be flexibly employed in our framework.

The sampled feature nodes are processed along two data paths: one for predicting the node-dependant dynamic affinities ${A'}^q \in \mathbb{R}^{H\times W\times K}$ and another path for dynamic filters $\vect{w}^q \in \mathbb{R}^{H\times W \times K \times G}$ where $K$ (\eg, $3 \times 3$) is the kernel size for the receiving node.
The matrix transformation $\vect{W}_{i,j}^{k,A}$ used to jointly predict the dynamic filters and affinities in Eq~\ref{equ:dynamicweights} is implemented by a $3 \times 3$ convolution layer.
Message $\vect{M}_q \in \mathbb{R}^{H\times W\times C}$ corresponding to the $q$-th sampling rate is
then scaled to perform a linear combination with the observed feature map $\vect{F}$, to produce a refined feature map $\vect{H}^{(1)}$ as output. 
To balance performance and efficiency, as in existing graph-based feature learning models~\cite{wang2018nonlocal,graph_reason, lin2015deeply, zhang2019dual}, we also perform $T = 1$ iteration of message updating. 

\subsection{Discussion}
Our approach is related to deformable convolution~\cite{dai2017deformable, zhu2018deformable} and Non-local~\cite{wang2018understanding}, but has several key differences:

A fundamental difference to deformable convolution is that it only learns the offset dependent on the input feature while the filter weights are fixed for all inputs. 
In contrast, our model learns the random walk, weight and affinity as all being dependent on the input.
This property makes our weights and affinities position-specific whereas deformable convolution shares the same weight across all convolution positions in the feature map.
Moreover, \cite{dai2017deformable,zhu2018deformable} only consider $3 \times 3$ local neighbours at each convolution position.
In contrast, our model, for each position, learns to sample a set of $K$ nodes (where $K \gg 9$) for message passing globally from the whole feature map. 
This allows our model to capture a larger receptive field than deformable convolution.

Whilst Non-local also learns to refine deep features, it uses a self-attention matrix to guide the message passing between each pair of feature nodes.
In contrast, our model learns to sample graph feature nodes to capture global feature information efficiently.
This dynamic sampling reduces computational overhead, whilst still being able to improve upon the accuracy of Non-local across multiple tasks as shown in the next section.

%% file: figures/framework.tex
\begin{figure*}
	\centering
	\vspace{-0.5\baselineskip}
	\includegraphics[width=0.85\linewidth]{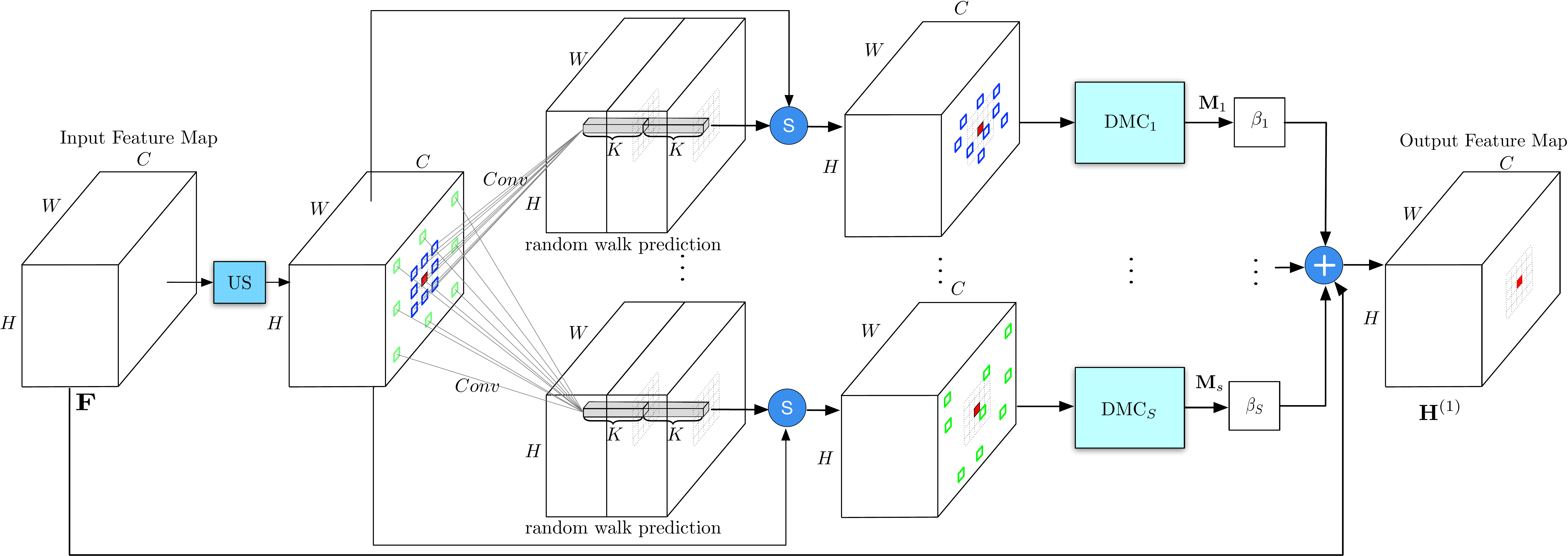}
	\caption{
	\small 
	Overview of our proposed dynamic graph message passing network (DGMN).
	The neighbourhood used to update the feature representation of each node (we show a single node with a red square) is predicted dynamically conditioned on each input.
	This is done by first uniformly sampling (denoted by ``US'') a set of $S$ neighbourhoods around each node.
	Each neighbourhood contains $K$ (\eg $3 \times 3$) sampled nodes.
	Here, the blue nodes were sampled with a low sampling rate, and the green ones with a high sampling rate.
	Walks are predicted (conditioned on the input) from these uniformly sampled nodes, denoted by the $\textcircled{s}$ symbol representing the random walk sampling operation described in Sec.~\ref{sec:method_sampling}.
	$\mathrm{DMC}_1, \cdots, \mathrm{DMC}_S$ and $\beta_1, \cdots, \beta_S$ denotes $S$ dynamic message calculation operations and $S$ message scaling parameters, respectively. 
	The DMC module is detailed in Figure~\ref{fig:dmc}.
	The symbol $\oplus$ indicates an element-wise addition operation.	
	}
	\label{fig:framework}
 	\vspace{-1\baselineskip}
\end{figure*}

%% file: figures/dmc.tex
\begin{figure*}
	\centering
	\includegraphics[width=0.85\linewidth]{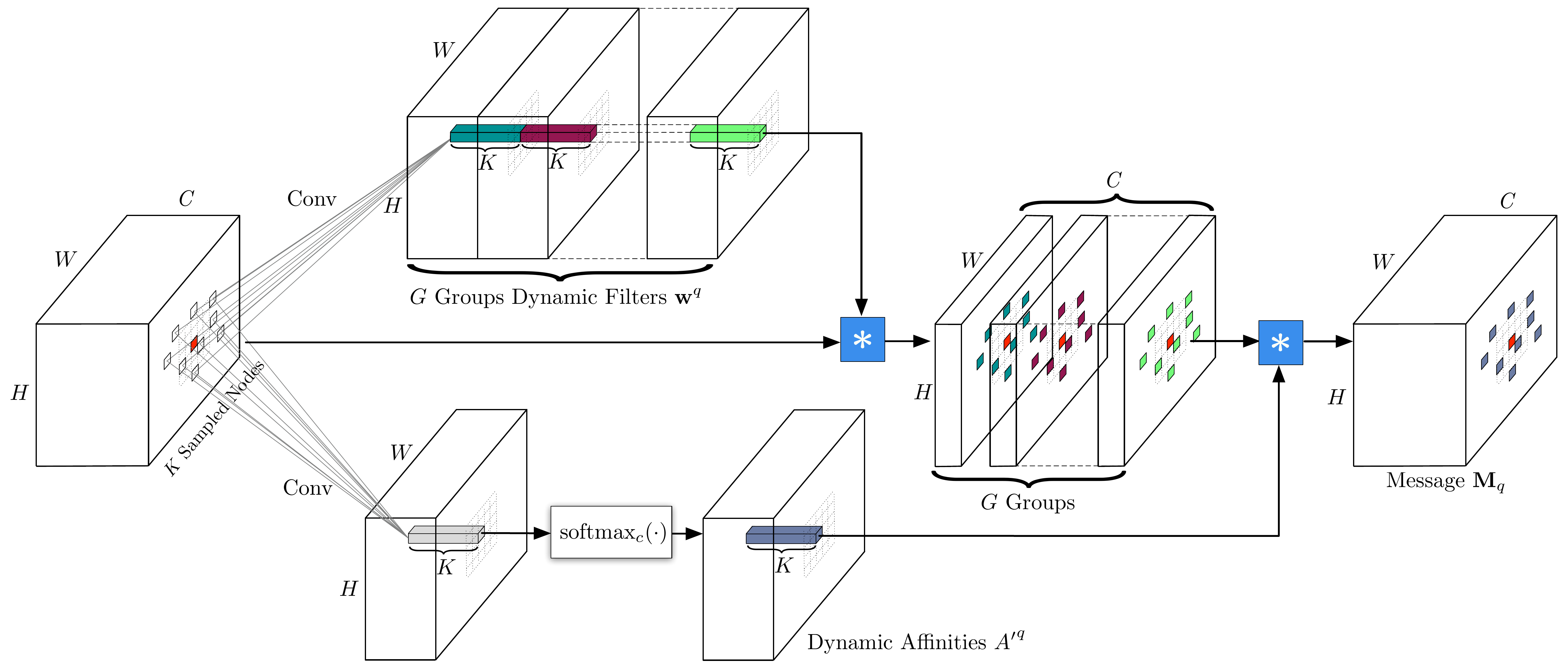}
 	\vspace{-0.6\baselineskip}
	\caption{\small Schematic illustration of the proposed dynamic message passing calculation (DMC) module. 
	The small red square indicates the receiving node whose message is calculated from its neighbourhood, \ie~the sampled $K$ (\eg $3 \times 3$) features nodes. 
	The module accepts a feature map as input and produces its corresponding message map. 
	The symbol $\ast$ denotes group convolution operation using the dynamically predicted and position specific group kernels and affinities.}
	\label{fig:dmc}
 	\vspace{-1\baselineskip}
\end{figure*}

%% file: 5-experiments.tex
\section{Experiments}
\label{sec:exp}
\input{figures/vis.tex}
\subsection{Experimental setup}
\label{sec:exp_setup}

\par\noindent\textbf{Tasks and datasets.}
We evaluate our proposed model on two challenging public benchmarks, \ie,~{Cityscapes}~\cite{Cityscapes} for semantic segmentation, and {COCO}~\cite{lin2014microsoft} for object detection and instance segmentation.
Both datasets have hidden test sets which are evaluated on a public evaluation server.
We follow the standard protocol and evaluation metrics used by these public benchmarks. 
More details can be found in supplementary material.

\par\noindent\textbf{Baseline models.}
For semantic segmentation on Cityscapes, our baseline is Dilated-FCN~\cite{yu2015multi} with a ResNet-101 backbone pretrained on ImageNet.
A randomly initialised $3 \times 3$ convolution layer, together with batch normalisation and ReLU is used after the backbone to produce a dimension-reduced feature map of 512 channels which is then fed into the final classifier.
For the task of object detection and instance segmentation on COCO, our baseline is Mask-RCNN~\cite{maskrcnn,massa2018mrcnn} with FPN and ResNet/ResNeXt~\cite{resnet,xie2017aggregated} as a backbone architecture.
Unless otherwise specified, we use a {single scale} and test with a {single model} for all experiments without using other complementary performance boosting ``tricks''. 
We train models on the COCO training set and test on the validation and test-dev sets.

Across all tasks and datasets, we consider Non-local networks~\cite{wang2018nonlocal} as an additional baseline. 
To have a direct comparison with the Deformable Convolution method~\cite{dai2017deformable,zhu2018deformable}, we consider two baselines: 
(i) ``deformable message passing'', which is a variant of our model using randomly initialised deformable convolutions for message calculation, but without using the proposed dynamic sampling and dynamic weights/affinities strategies, and 
(ii) the original deformable method which replaces the convolutional operations as in~\cite{dai2017deformable,zhu2018deformable}.

\par\noindent\textbf{Implementation details.} 
For our experiments on Cityscapes, our DGMN module is randomly initialised and inserted between the $3 \times 3$ convolution layer and the final classifier.
For the experiments on COCO, we insert one or multiple randomly initialised DGMN modules into the backbone for deploying our approach for feature learning.
Our models and baselines for all of our COCO experiments are trained with the typical ``$1\mathrm{x}$'' training settings from the public Mask R-CNN benchmark~\cite{massa2018mrcnn}.

When predicting dynamic filter weights, we used the grouping parameter $G=4$. 
For our experiments on Cityscapes, the sample rates are set to $\varphi = \{1, 6, 12, 24, 36\}$.
For experiments on COCO, we use smaller sampling rates of $\varphi = \{1, 4, 8, 12\}$.	
The effect of this hyperparameter 
and additional implementation details are described in the supplementary material.

\subsection{Model analysis}
\label{sec:exp_ablation}

To demonstrate the effectiveness of the proposed components of our model, we conduct ablation studies of: 
(i) DGMN w/ DA, which adds the dynamic affinity (DA) strategy onto the DGMN base model; 
(ii) DGMN w/ DA+DW, which further adds the proposed dynamic weights (DW) prediction; 
(iii) DGMN w/ DA+DW+DS, which is our full model with the dynamic sampling (DS) scheme added upon DGMN w/ DA+DW.
Note that DGMN Base was described in Sec.~\ref{sec:method_mpgraph}, DS in Sec.~\ref{sec:method_sampling}, and DA and DW in \ref{sec:method_dynamic_weights}.

\input{tables/cs_val.tex}

\par\noindent\textbf{Effectiveness of the dynamic sampling strategy.} 
Table~\ref{tab:ablation_study_cityscapes} shows quantitative results of semantic segmentation on the Cityscapes validation set. 
DGMN w/ DA+DW+DS clearly outperforms DGMN w/ DA+DW on the challenging segmentation task, meaning that the feature-conditioned adaptive sampling based on learned random walks is more effective compared to a spatial uniform sampling strategy when selecting nodes. 
More importantly, all variants of our module for both semantic segmentation and object detection that use dynamic sampling (Tab.~\ref{tab:ablation_study_cityscapes} and Tab.~\ref{tab:ablation_study_coco}) achieve higher performance than a fully-connected model (\ie,~Non-local~\cite{wang2018nonlocal}) with substantially fewer FLOPs. This emphasises the performance benefits of our dynamic graph sampling model.
Visualisations of the nodes dynamically sampled by our model are shown in Fig.~\ref{fig:offset}.

\input{tables/ablation_study_coco}
\input{figures/apcurve.tex}

\par\noindent\textbf{Effectiveness of joint learning the dynamic filters and affinities.} 
As shown in Tab.~\ref{tab:ablation_study_cityscapes}, DGMN w/ DA is 1.5 points better than Dilated FCN baseline with only a slight increase in FLOPs and parameters, showing the benefit of using predicted dynamic affinities for reweighting the messages in message passing. 
By further employing the estimated dynamic filter weights for message calculation, the performance increases substantially from a mIoU of 76.5\% to 79.1\%, which is almost the same as the 79.2\% of the Non-local model~\cite{wang2018nonlocal}.
Crucially, our approach only uses 9.4\% of the FLOPs and 25.3\% of the parameters compared to Non-local.
These results clearly demonstrate our motivation of jointly learning the dynamic filters and dynamic affinities from sampled graph nodes. 

\par\noindent\textbf{Comparison with other baselines.} 
For semantic segmentation on Cityscapes, we clearly outperform the ASPP module of Deeplab v3 \cite{deeplabv3} which also increases the receptive field by using multiple dilation rates.
Notably, we improve upon the most related method, Non-local~\cite{wang2018nonlocal} which models a fully-connected graph, whilst using only 33\% of the FLOPs of~\cite{wang2018nonlocal}.
This suggests that a fully-connected graph models redundant information, and further confirms the performance and efficiency of our model.

For fair comparison with Non-local ~\cite{wang2018nonlocal} as well as other alternatives on COCO, we insert one randomly initialised DGMN module right before the last residual block of \textit{res4}~\cite{wang2018nonlocal} (Tab.~\ref{tab:ablation_study_coco}).
We also compare to GCNet~\cite{cao2019gcnet} and CCNet~\cite{huang2018ccnet} which both aim to reduce the complexity of the fully connected Non-local model.
Our proposed DGMN model substantially improves upon these strong baselines and alternatives. 
Figure~\ref{fig:apcurve} further shows the validation curves of our method and different baselines using the standard $\text{AP}^{\text{Box}}$ and $\text{AP}^{\text{Mask}}$ measures for semantic and instance segmentation respectively.
Our method is consistently better than the Non-local and Mask R-CNN baselines throughout training.

\input{figures/qualitative.tex}

\par\noindent\textbf{Effectiveness of multiple DGMN modules.} 
We further show the effectiveness of our approach for representation learning by inserting multiple of our DGMN modules into the ResNet-50 backbone. 
Specifically, we add our full DGMN module \textit{after} all $3 \times 3$ conv layers in \textit{res5} which we denote as ``C5''.
The second part of Tab.~\ref{tab:ablation_study_coco} shows that our model significantly improves upon the Mask R-CNN baseline 
with the improvements of 2.4 points for the AP$^{\mathrm{box}}$ on object detection, and 1.6 points for the AP$^{\mathrm{mask}}$ on instance segmentation.
Furthermore, when we insert the GCNet module~\cite{cao2019gcnet} in the same locations for comparison, our model achieves better performance too. 
A straightforward comparison to the deformable method of \cite{zhu2018deformable} is the deformable message passing baseline in which we disable the proposed dynamic sampling and the dynamic weights and affinities learning strategies. 
In Tab.~\ref{tab:ablation_study_coco},
our model significantly improves upon the deformable message passing method, which is a direct evidence of the effectiveness of jointly modeling the dynamic sampling and dynamic filters and affinities for feature learning.
Furthermore, we also consider inserting the improved Deformable Conv method~\cite{zhu2018deformable} in C5, which is complementary to our model since it is plugged \textit{before} $3 \times 3$ layers to replace the convolution operations.
Our approach also achieves better performance than it.

\input{tables/cs_test}
\input{tables/coco_test}

\subsection{Comparison to State-of-the-art}
\label{sec:exp_sota}
\textbf{Performance on Cityscapes test set.}
Table~\ref{tab:city_test} compares our approach with state-of-the-art methods on Cityscapes.
Note that all methods are trained using only the fine annotations and evaluated on the public evaluation server as test-set annotations are withheld from the public.
As shown in the table, DGMN (ours) achieves an mIoU of 81.6\%, surpassing all previous works. 
Among competing methods, GloRe~\cite{graph_reason}, Non-local~\cite{wang2018nonlocal}, CCNet~\cite{huang2018ccnet} and DANet~\cite{DAnet} are the most related to us as they all based on graph neural network modules. 
Note that we followed common practice and employed several complementary strategies used in semantic segmentation to boost performance,
including Online Hard Example Mining (OHEM)~\cite{DAnet}, Multi-Grid~\cite{deeplabv3} and Multi-Scale (MS) ensembling~\cite{pspnet}. 
The contribution of each strategy on the final performance is reported in the supplementary. 

\par\noindent\textbf{Performance on COCO 2017 test set.} 
Table~\ref{tab:test_coco} presents our results on the COCO test-dev set, where we inserted our module on multiple backbones. 
By inserting DGMN into all layers of C4 and C5, we substantially improve the performance of Mask R-CNN, observing a gain of of 3.0 and 2.2 points on the AP$^{\mathrm{box}}$ and the AP$^{\mathrm{mask}}$ of object detection and instance segmentation respectively.
We observe similar improvements when using the ResNet-101 or ResNeXt-101 backbones as well, showing that our proposed DGMN module generalises to multiple backbone architectures.

Note that the Mask-RCNN baseline with ResNet-101 has 63.1M parameters and 354 GFLOPs.
Our DGMN (C4, C5) model with ResNet-50 outperforms it whilst having only 51.1M parameters and 297.1 GFLOPs.
This further shows that the improvements from our method are due to the model design, and not only the increased parameters and computation.
Further comparisons to the state-of-the-art are included in the supplementary.

%% file: figures/vis.tex
\begin{figure*}[!t]
	\def \imheight {2.50cm}
	\centering
	\includegraphics[height=\imheight]{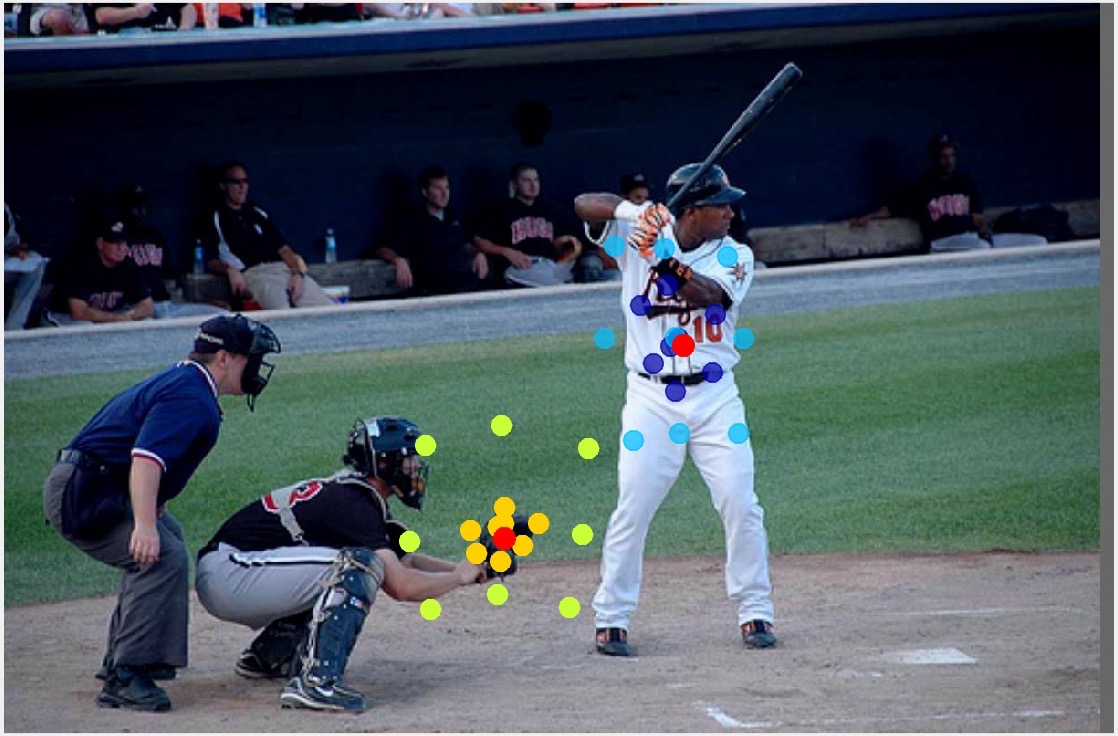}
	\includegraphics[height=\imheight]{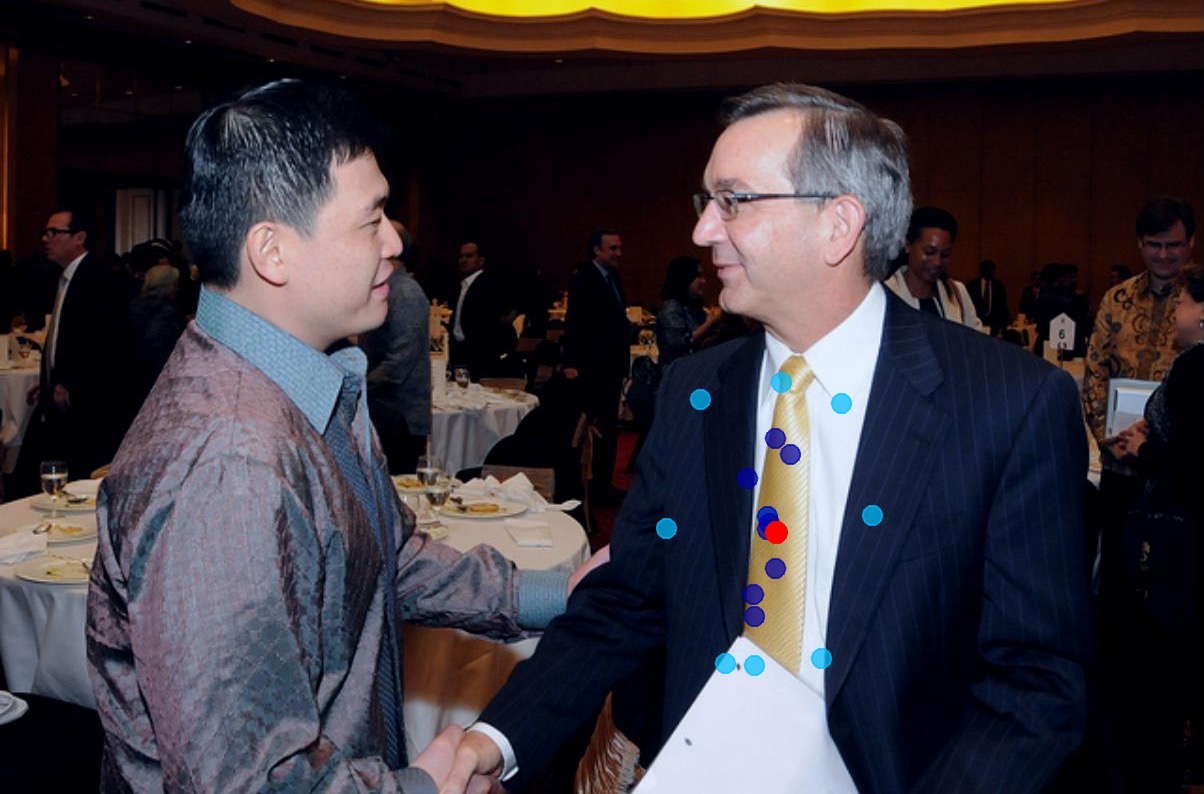}
	\includegraphics[height=\imheight]{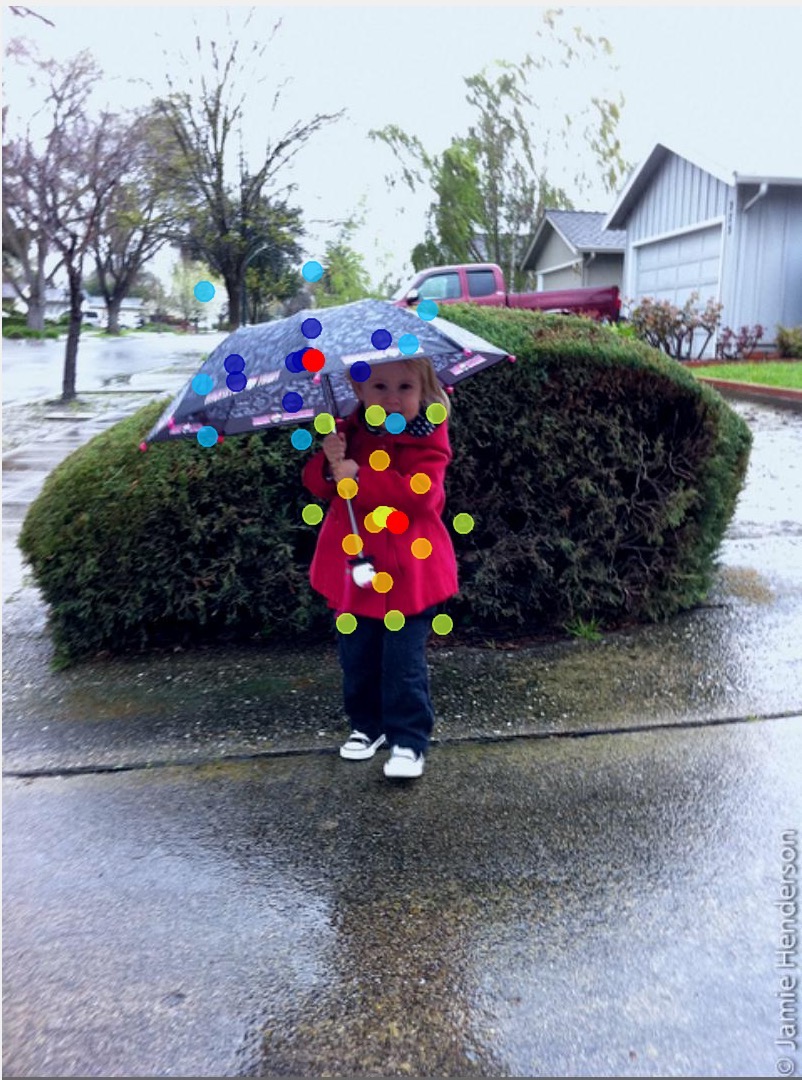} 
	\includegraphics[height=\imheight]{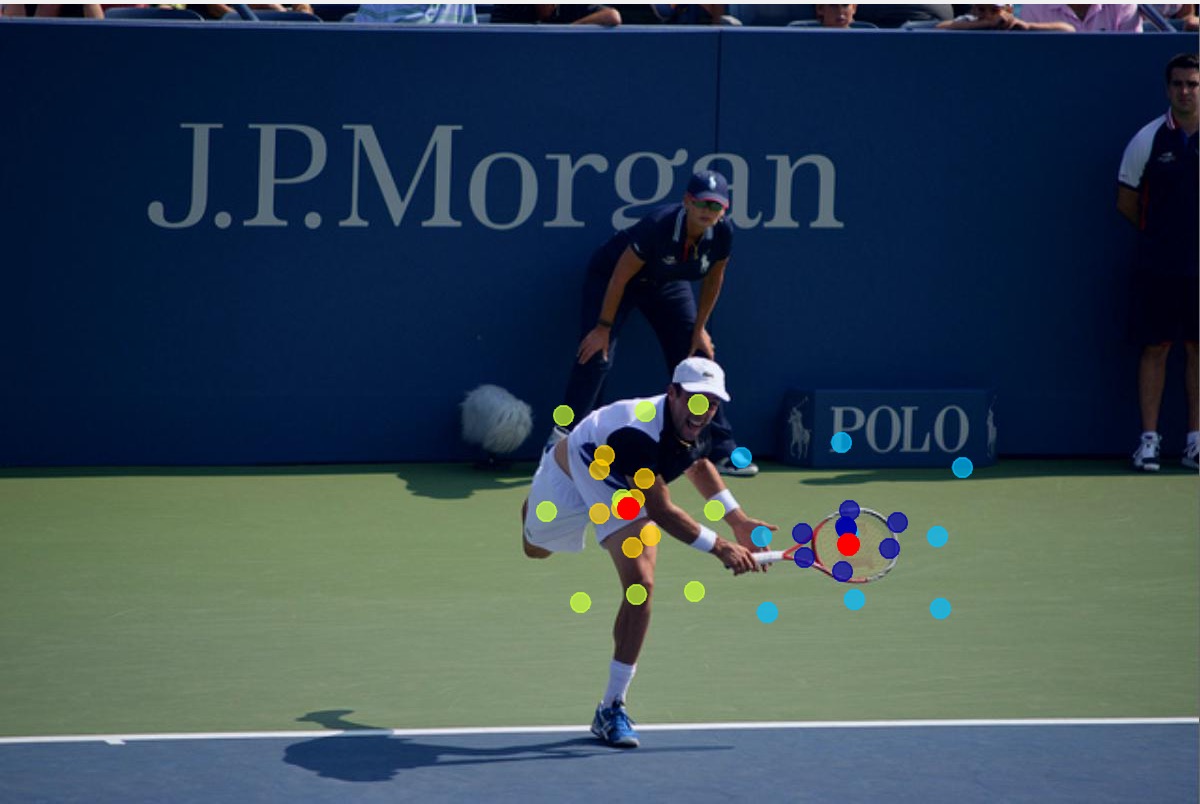}
	\includegraphics[height=\imheight]{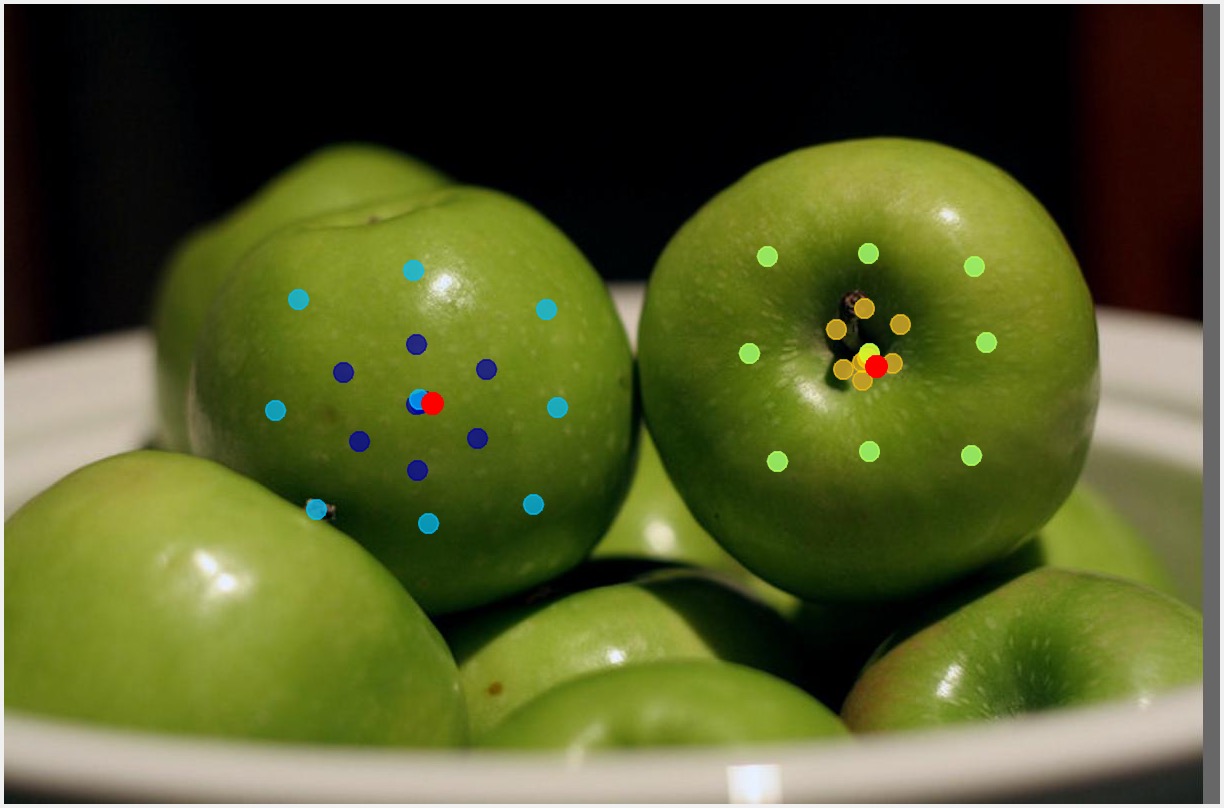}
	\caption{\small Visualisation of the nodes sampled via \textit{learning} the random walks with our network (trained for instance segmentation on COCO).  
	The red point indicates a receiving node $i$. 
	Different colour families (\ie~yellow and blue) indicate the learned position specific weights and affinities of the sampled nodes.
	Different colours in the same family show the sampled nodes with different sampling rates for the same receiving node. 
	}
	\label{fig:offset}
 	\vspace{-1.5\baselineskip}
\end{figure*}

%% file: tables/cs_val.tex
\begin{table}[t]
	\centering
	{\tablestyle{2pt}{1.2}
		\begin{tabular}{ l|x{34}x{34}x{34} }
		&  mIoU (\%)  &   Params &   FLOPs \\
			\shline
			Dilated FCN~\cite{yu2015multi}  & 75.0 &  -- &  --\\
            \hline
             + Deformable~\cite{zhu2018deformable} & 78.2 & +1.31M  &+12.34G \\
             + ASPP~\cite{deeplabv3} &78.9&  +4.42M & +38.45G \\
             + Non-local~\cite{wang2018nonlocal}     &  79.0 & +2.88M & +73.33G\\
            \hline
             + DGMN w/ DA   & 76.5   & +0.57M & +5.32G\\
             + DGMN w/ DA+DW   & 79.1   & +0.73M & +6.88G\\
             + DGMN w/ DA+DW+DS   & \textbf{80.4}   & +2.61M & +24.55G\\
		\end{tabular}
	}
	\caption{Ablation study on the Cityscapes validation set for semantic segmentation. All models have a ResNet-101 backbone and are evaluated at a single scale.}
	\label{tab:ablation_study_cityscapes}
	\vspace{-\baselineskip}
\end{table}

%% file: tables/ablation_study_coco.tex
\begin{table}
	\centering
	\tablestyle{1.2pt}{1}
    \begin{tabular}{l|x{21}x{21}x{21}|x{21}x{21}x{21}}
     &\scriptsize $\mathrm{AP^{b}}$ &\scriptsize $\mathrm{AP^{b}_{50}}$ &\scriptsize $\mathrm{AP^{b}_{75}}$ &\scriptsize $\mathrm{AP^{m}}$ &\scriptsize $\mathrm{AP^{m}_{50}}$ &\scriptsize $\mathrm{AP^{m}_{75}}$  \\
    \shline
    \scriptsize Mask R-CNN baseline   &\scriptsize 37.8&\scriptsize 59.1 &\scriptsize 41.4 &\scriptsize 34.4  &\scriptsize 55.8 &\scriptsize 36.6 \\
    \hline
    \scriptsize + GCNet~\cite{cao2019gcnet} &\scriptsize 38.1 &\scriptsize  60.0 &\scriptsize41.2 &\scriptsize 34.9 &\scriptsize 56.5 &\scriptsize37.2  \\
    \scriptsize + Deformable Message Passing &\scriptsize 38.7 & \scriptsize 60.4 &\scriptsize42.4 &\scriptsize 35.0 &\scriptsize 56.9 &\scriptsize37.4  \\
    \scriptsize + Non-local~\cite{wang2018nonlocal}    
    &\scriptsize 39.0 &\scriptsize 61.1 &\scriptsize 41.9 &\scriptsize 35.5  &\scriptsize 58.0 &\scriptsize 37.4 \\
    \scriptsize + CCNet~\cite{huang2018ccnet}    &\scriptsize39.3 &\scriptsize- &\scriptsize- &\scriptsize36.1  &\scriptsize- &\scriptsize- \\

    \scriptsize+ DGMN &\scriptsize \bf 39.5 & \scriptsize \bf 61.0 &\scriptsize \bf 43.3&\scriptsize \bf 35.7&\scriptsize \bf 58.0 &\scriptsize \bf 37.9 \\ 
    \hline
    \scriptsize + GCNet (C5) ~\cite{cao2019gcnet}    &\scriptsize38.7 &\scriptsize61.1 &\scriptsize41.7 &\scriptsize35.2 &\scriptsize57.4 &\scriptsize37.4 \\ 
    \scriptsize + Deformable (C5) ~\cite{zhu2018deformable}     &\scriptsize39.9&\scriptsize- &\scriptsize- &\scriptsize34.9  &\scriptsize- &\scriptsize- \\ 
    \scriptsize+ DGMN (C5)  &\scriptsize\textbf{40.2}&\scriptsize\textbf{62.0} &\scriptsize\textbf{43.4} &\scriptsize\textbf{36.0}  &\scriptsize\textbf{58.3} &\scriptsize\textbf{38.2} \\ 
    \multicolumn{7}{c}{~}\\
    \end{tabular}
    \vspace{-1.5\baselineskip}
    \caption{ 
    \small Quantitative results of different models on the COCO 2017 validation set for object detection ($\mathrm{AP^{b}}$) and instance segmentation ($\mathrm{AP^{m}}$). 
    C5 denotes inserting DGMN after all $3 \times 3$ convolutional layers in \emph{res5}.
    All methods are based on the Mask R-CNN with ResNet-50 as backbone. 
    }
    \label{tab:ablation_study_coco}
    \vspace{-0.5\baselineskip}
\end{table}

%% file: figures/apcurve.tex
\begin{figure}[t]
	\centering
	\includegraphics[width=0.86\linewidth]{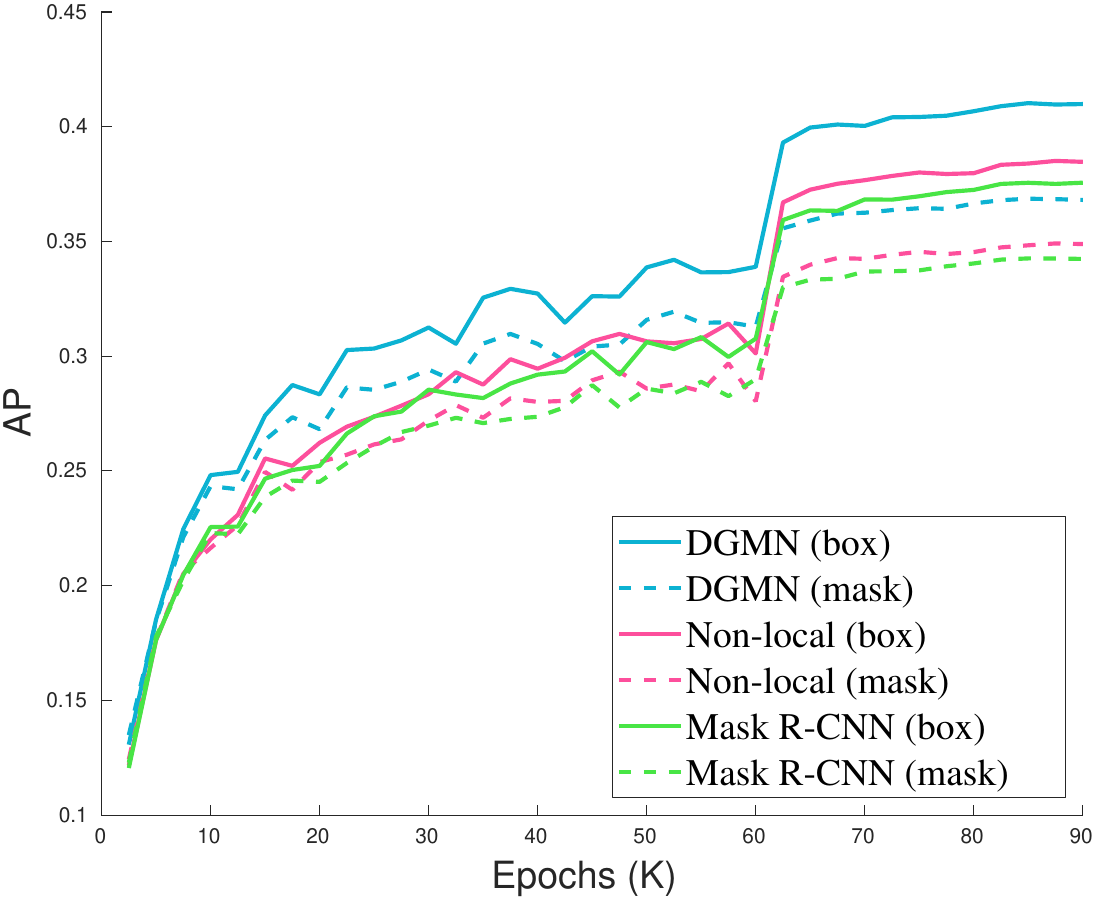}
	\vspace{-.3cm}
	\captionof{figure}{
	\small Validation curves of $\mathrm{AP^{box}}$ and $\mathrm{AP^{mask}}$ on COCO for Mask-RCNN baseline, Non-local and the proposed DGMN.
	The number of training epochs is 90K.
	}
	\label{fig:apcurve}
	\vspace{-1.5\baselineskip}
\end{figure}

%% file: figures/qualitative.tex
\begin{figure*}[!t]
	\centering
	\def \imwidth {4cm}
	\small
	\begin{tabularx}{\linewidth}{YYYY}
	\includegraphics[width=\imwidth]{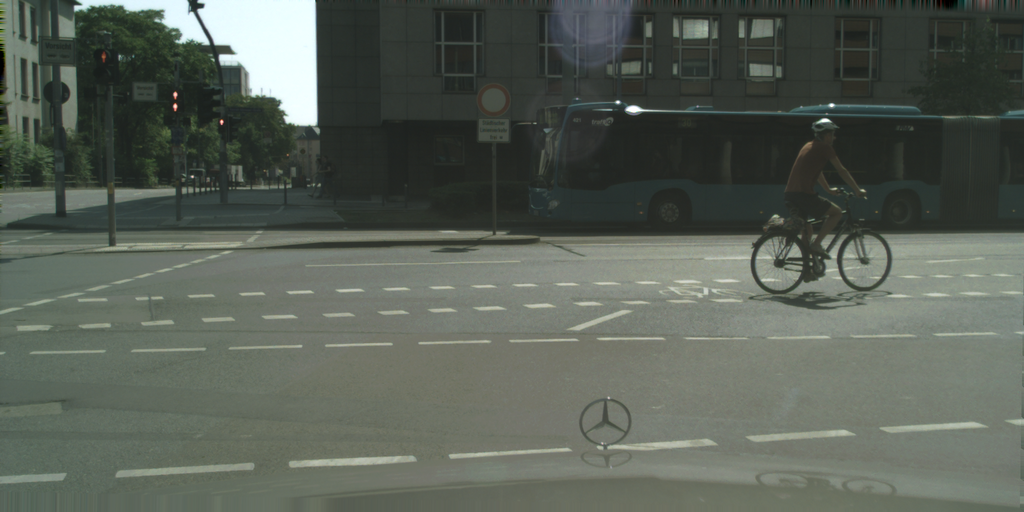} &
	\includegraphics[width=\imwidth]{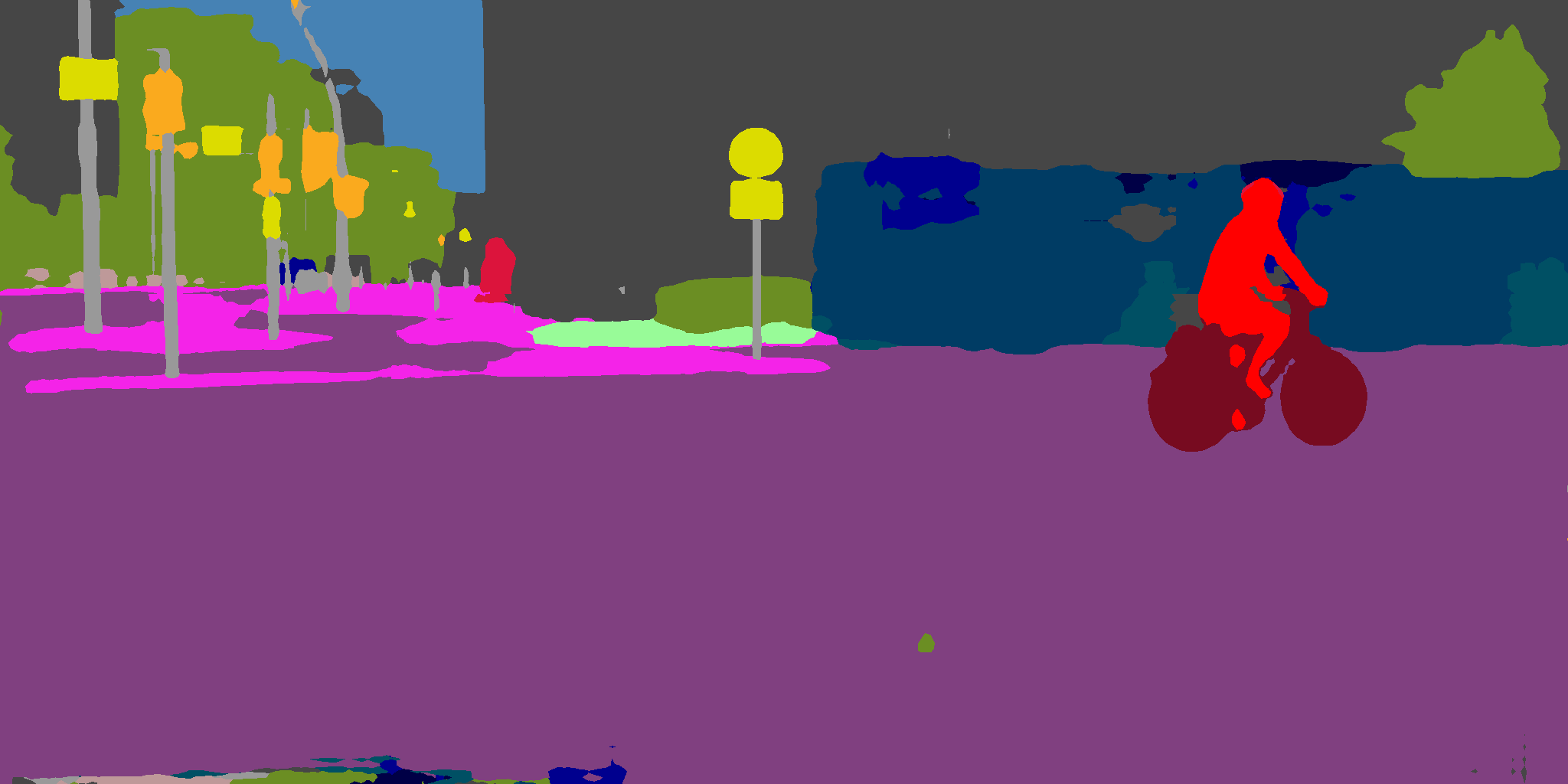} &
	\includegraphics[width=\imwidth]{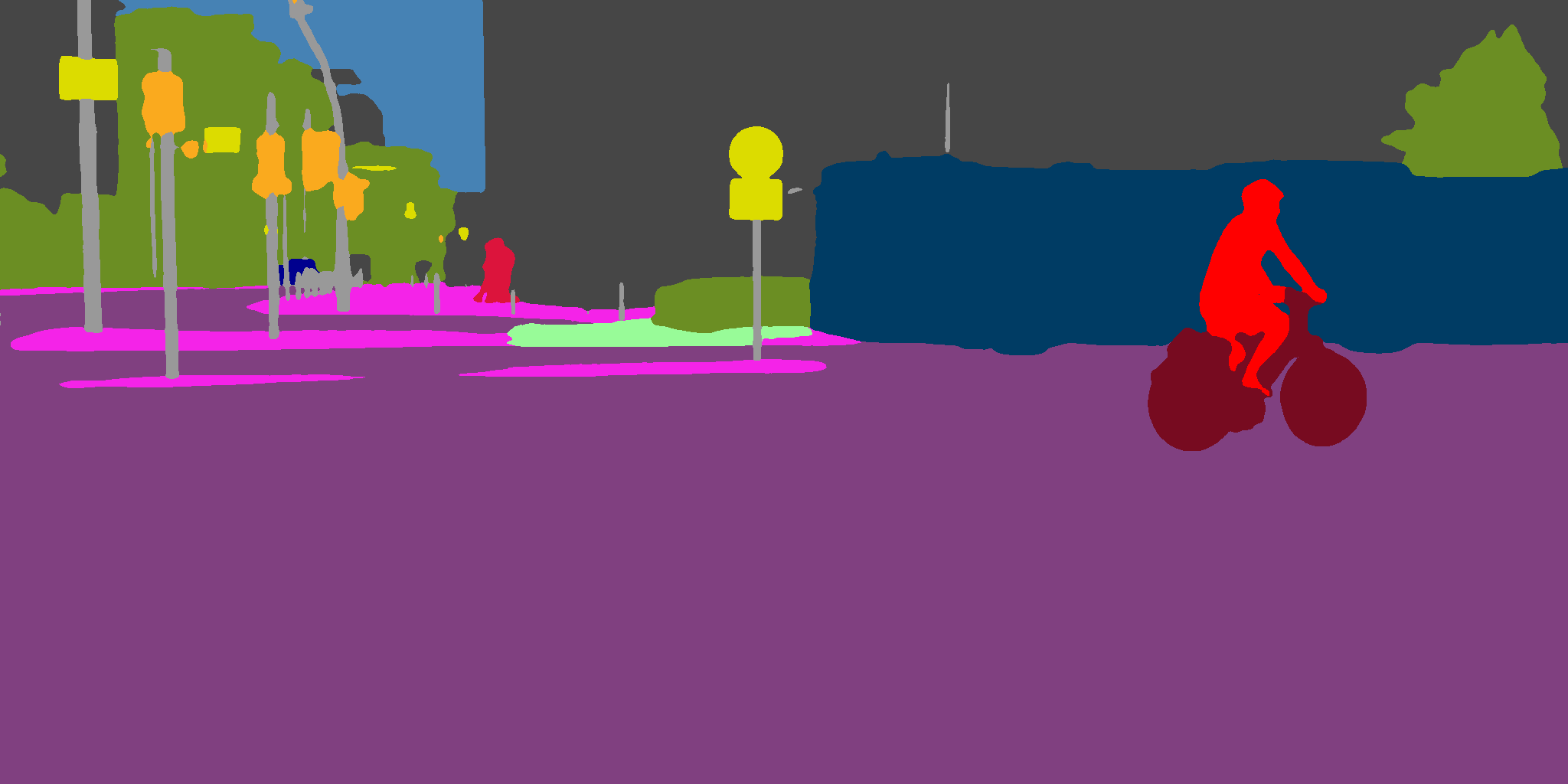} &
	\includegraphics[width=\imwidth]{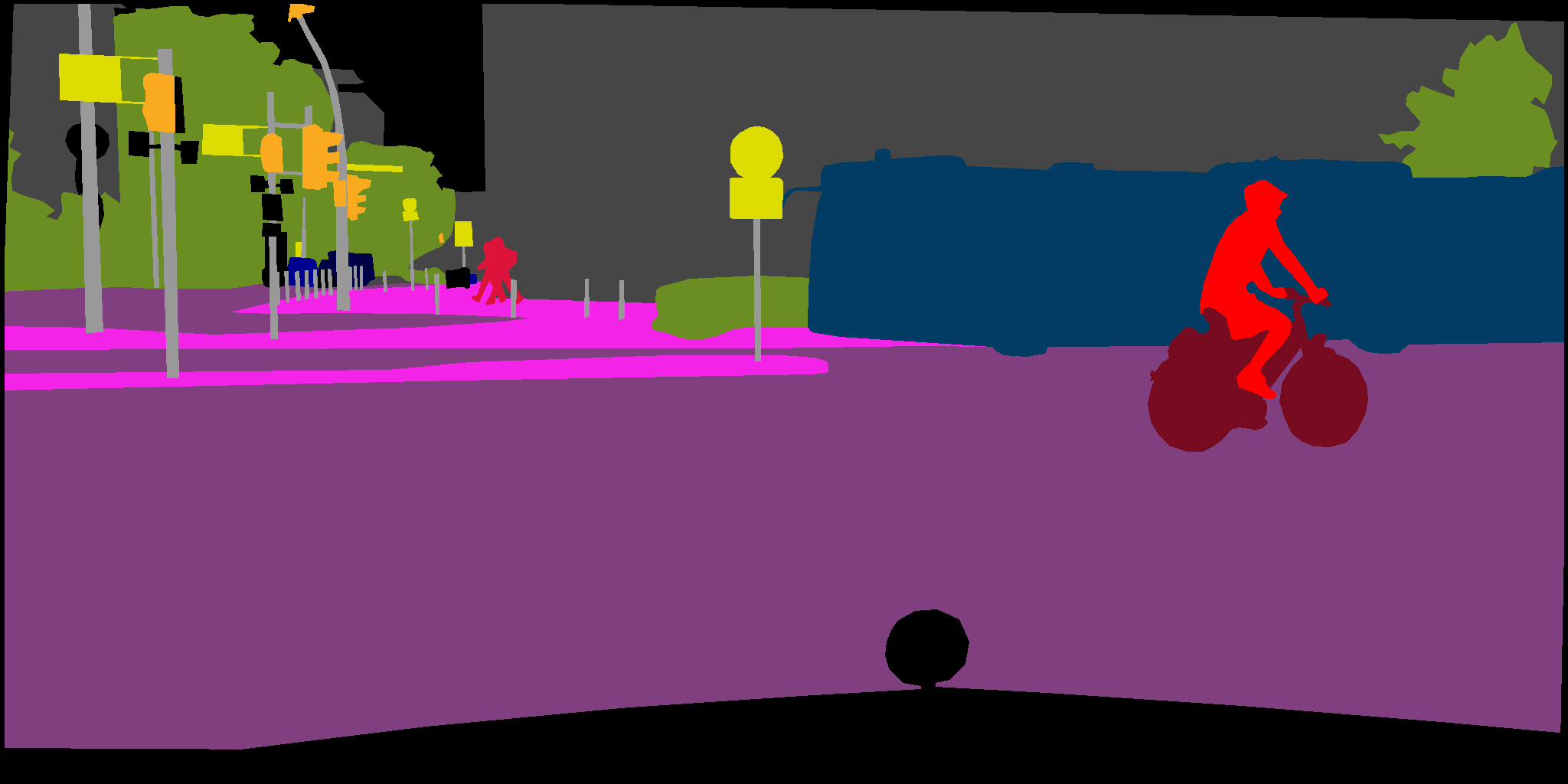}
	\\
	Input image & Dilated FCN & DGMN (ours) & Ground truth \\
	\includegraphics[width=\imwidth]{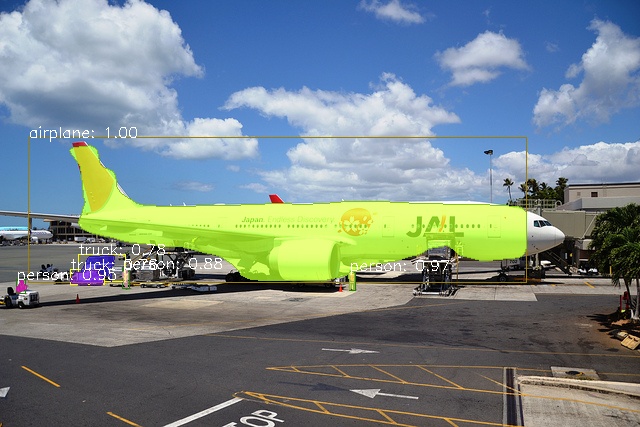} &
	\includegraphics[width=\imwidth]{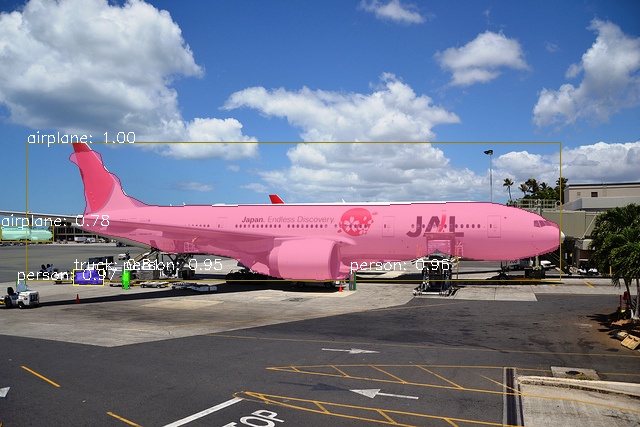} &
	\includegraphics[width=\imwidth]{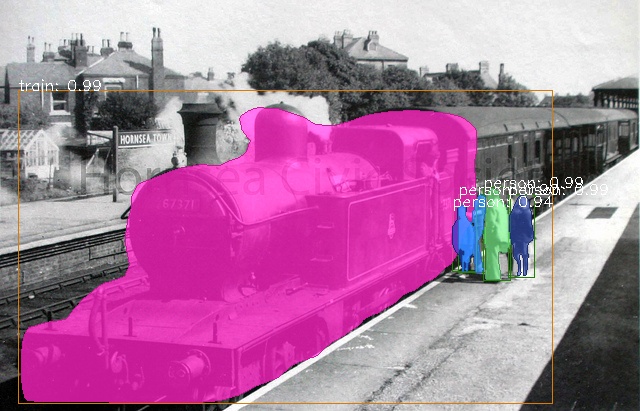} &
	\includegraphics[width=\imwidth]{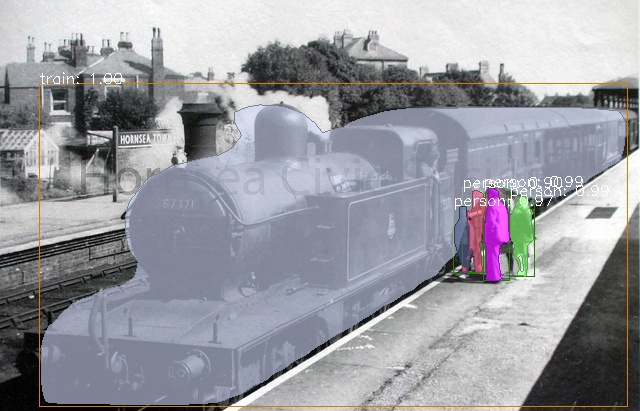} \\
	Mask R-CNN & DGMN (ours) & Mask R-CNN & DGMN (ours) \\
	\end{tabularx}	
	\vspace{-\baselineskip}
	\caption{\small Qualitative examples of our results for  semantic segmentation on Cityscapes (first row), and object detection and instance segmentation on COCO (second row)
	}
	\label{fig:qualitative}
  	\vspace{-1.5\baselineskip}
\end{figure*}

%% file: tables/cs_test.tex
\begin{table}[t]
	\centering
	\label{tab:city_test}{
		\tablestyle{1pt}{1}
		\begin{tabular}{l |x{42}|x{42}}
			&\scriptsize Backbone &\scriptsize mIoU (\%) \\
			\shline
			\scriptsize PSPNet~\cite{pspnet}   &\scriptsize ResNet 101  &\scriptsize 78.4\\
            \scriptsize PSANet~\cite{psanet} &\scriptsize ResNet 101  &\scriptsize 80.1   \\
            \scriptsize DenseASPP~\cite{denseaspp} &\scriptsize DenseNet 161   &\scriptsize 80.6   \\
            \hline
            \scriptsize GloRe~\cite{graph_reason} &\scriptsize ResNet 101  &\scriptsize 80.9  \\
            \scriptsize Non-local~\cite{wang2018nonlocal} &\scriptsize ResNet 101 &\scriptsize 81.2 \\
            \scriptsize CCNet~\cite{huang2018ccnet} &\scriptsize ResNet 101 &\scriptsize 81.4 \\
            \scriptsize DANet~\cite{DAnet} &\scriptsize ResNet 101     &\scriptsize 81.5   \\
            \scriptsize DGMN (Ours)  &\scriptsize ResNet 101   &\scriptsize \bf 81.6   \\ 
		\end{tabular}}
	\vspace{-.2cm}
	\caption{
	Comparison to state-of-the-art for semantic segmentation on Cityscapes.
	All methods are trained with the finely-annotated data from the training and validation sets.
	}
	\label{tab:city_test}
	\vspace{-0.5\baselineskip}
\end{table}

%% file: tables/coco_test.tex
\begin{table}[t]
\centering
\scalebox{0.75}{
\tablestyle{1pt}{1}
\begin{tabular}{l|x{40}|x{32}x{32}x{32}|x{32}x{32}x{32}}
&\scriptsize Backbone &\scriptsize $\mathrm{AP^{box}}$ &\scriptsize $\mathrm{AP^{box}_{50}}$ &\scriptsize $\mathrm{AP^{box}_{75}}$ &\scriptsize $\mathrm{AP^{mask}}$ &\scriptsize $\mathrm{AP^{mask}_{50}}$ &\scriptsize $\mathrm{AP^{mask}_{75}}$ \\
\shline
\scriptsize Mask R-CNN baseline &\scriptsize \multirow{3}{*}{ResNet 50} &\scriptsize 38.0 &\scriptsize59.7 &\scriptsize41.5 &\scriptsize 34.6 &\scriptsize56.5 &\scriptsize36.6\\
\scriptsize+ DGMN (C5) & &\scriptsize40.2  &\scriptsize 62.5 &\scriptsize 43.9 &\scriptsize 36.2 &\scriptsize 59.1 &\scriptsize  38.4\\
\scriptsize+ DGMN (C4, C5)& &\scriptsize \bf41.0 &\scriptsize \bf63.2 &\scriptsize \bf44.9 &\scriptsize \bf36.8 &\scriptsize \bf59.8 &\scriptsize \bf39.1 \\
\hline
\scriptsize Mask R-CNN baseline & \scriptsize  \multirow{2}{*}{ResNet 101}   &\scriptsize40.2 &\scriptsize61.9 &\scriptsize44.0 &\scriptsize36.2  &\scriptsize58.6 &\scriptsize38.4\\
\scriptsize+ DGMN (C5)& &\scriptsize \bf 41.9  &\scriptsize \bf 64.1&\scriptsize \bf 45.9 &\scriptsize \bf 37.6 &\scriptsize \bf 60.9 &\scriptsize \bf 40.0 \\
\hline
\scriptsize Mask R-CNN baseline&\scriptsize \multirow{2}{*}{ResNeXt 101}  &\scriptsize 42.6 &\scriptsize 64.9 &\scriptsize 46.6 &\scriptsize 38.3  &\scriptsize 61.6 &\scriptsize 40.8\\
\scriptsize+ DGMN (C5)& &\scriptsize \bf 44.3   &\scriptsize \bf 66.8 &\scriptsize \bf 48.4  &\scriptsize \bf 39.5 &\scriptsize \bf 63.3 &\scriptsize \bf 42.1 \\
\end{tabular}
}
\vspace{-0.5\baselineskip}
\caption{\small Quantitative results via plugging our DGMN module on different backbones on the COCO 2017 \texttt{test-dev} set for object detection ($\mathrm{AP^{box}}$) and instance segmentation ($\mathrm{AP^{mask}}$).}
\label{tab:test_coco}
\vspace{-1\baselineskip}
\end{table}

%% file: 6-conclusion.tex
\section{Conclusion}
\label{sec:conclusion}

We proposed Dynamic Graph Message Passing Networks, a novel graph neural network module that dynamically determines the graph structure for each input.
It learns dynamic sampling of a small set of relevant neighbours for each node, and also predicts the weights and affinities dependant on the feature nodes to propagate information through this sampled neighbourhood.
This formulation significantly reduces the computational cost of static, fully-connected graphs such as Non-local ~\cite{wang2018nonlocal} which contain many redundancies.
This is demonstrated by the fact that we are able to clearly improve upon the accuracy of Non-local, and several state-of-art baselines, on three complex scene understanding problems.

%% file: acknowledge.tex
\vspace{0.5\baselineskip}
\small{
\par\noindent\textbf{Acknowledgements.}{~We thank Professor Andrew Zisserman for valuable discussions.
This work was supported by the EPSRC grant Seebibyte EP/M013774/1, 
ERC grant ERC-2012-AdG 321162-HELIOS and 
EPSRC/MURI grant EP/N019474/1.
We would also like to acknowledge the Royal Academy of Engineering.
}
}

%% file: 7-supp.tex
\appendix
\section{Additional experiments}
\label{sec:appendix}

In this supplementary material, we report additional qualitative results of our approach (Sec.~\ref{sec:app_qualitative}), additional details about the experiments in our paper (Sec.~\ref{sec:app_exp_details}), and also conduct further ablation studies (Sec.~\ref{sec:app_ablation}).

\subsection{Qualitative results}
\label{sec:app_qualitative}

Figure~\ref{fig:cs_results} shows qualitative results for semantic segmentation (on Cityscapes) while 
Figure \ref{fig:coco} and \ref{fig:coco2} show qualitative results for instance segmentation (on COCO).

\subsection{Additional experimental details}
\label{sec:app_exp_details}

\subsubsection{Datasets}
\label{sec:app_datasets}

\par\noindent\textbf{Cityscapes:} Cityscapes~\cite{Cityscapes} has densely annotated semantic labels for 19 categories in urban road scenes, and contains a total of 5000 finely annotated images, divided into 2975, 500, and 1525 images for training, validation and testing respectively. 
We do not use the coarsely annotated data in our experiments. 
The images of this dataset have a high resolution of $1024 \times 2048$. 
Following the standard evaluation protocol~\cite{Cityscapes}, the metric of mean Intersection over Union (mIoU) averaged over all classes is reported. 

\vspace{\baselineskip}

\par\noindent\textbf{COCO:} COCO 2017~\cite{lin2014coco} consists of 80 object classes with a training set of 118,000 images, a validation set of 5000 images, and a test set of 2000 images. 
We follow the standard COCO evaluation metrics \cite{lin2014microsoft} to evaluate the performance of object detection and instance segmentation, employing the metric of mean average-precision (mAP) at different box and mask IoUs respectively.

\subsubsection{Semantic segmentation on Cityscapes}
\label{sec:app_implementation_cityscapes}

For the semantic segmentation task on Cityscapes, we follow \cite{pspnet} and use a polynomial learning rate decay with an initial learning rate of 0.01.
The momentum and the weight decay are set to 0.9 and 0.0001 respectively. 
We use 4 Nvidia V100 GPUs, batch size 8 and train for $40 000$ iterations from an ImageNet-pretrained model.
For data augmentation, random cropping with a crop size of 769 and random mirror-flipping are applied on-the-fly during training. 
Note that following common practice \cite{pspnet,ocnet,psanet,rota2018place} we used synchronised batch normalisation for better estimation of the batch statistics for the experiments on Cityscapes.
When predicting dynamic filter weights, we use the grouping parameter $G=4$. 
For the experiments on Cityscapes, we use a set of the sampling rates of $\varphi = \{1, 6, 12, 24, 36\}$. 

\subsubsection{Object detection and instance segmentation on COCO}
\label{sec:app_implementation_coco}

Our models and all baselines are trained with the typical ``$1\mathrm{x}$'' training settings from the public Mask R-CNN benchmark~\cite{massa2018mrcnn} for all experiments on COCO.
More specificially, 
the backbone parameters of all the models in the experiments are pretrained on ImageNet classification. 
The input images are resized such that their shorter side is of 800 pixels and the longer side is limited to 1333 pixels.
The batch size is set to 16. 
The initial learning rate is set to 0.02 with a decrease by a factor of 0.1 after $60 000$ and $80 000$ iterations, and finally terminates at $90 000$ iterations. 
Following~\cite{massa2018mrcnn, goyal2017accurate}, the training warm-up is employed by using a smaller learning rate of 0.02 $\times$ 0.3 for the first 500 iterations of training. 
All the batch normalisation layers in the backbone are ``frozen'' during fine-tuning on COCO. 

When predicting dynamic filter weights, we use the grouping parameter $G=4$. 
For the experiments on COCO, a set of the sampling rates of $\varphi = \{1, 4, 8, 12\}$ is considered.
We train models on only the COCO training set and test on the validation set and test-dev set.

\subsection{Additional ablation studies}
\label{sec:app_ablation}

\par\noindent\textbf{Effectiveness of different training and inference strategies.} 
When evaluating models for the Cityscapes test set, we followed common practice and employed several complementary strategies used to improve performance in semantic segmentation, including Online Hard Example Mining (OHEM)~\cite{ohem,pohlen2017full,li2017holistic,ocnet,bisenet}, Multi-Grid~\cite{deeplabv3,DAnet,graph_reason} and Multi-Scale (MS) ensembling~\cite{deeplabv1,deeplabv3p,pspnet,ocnet,boxsup}. 
The contribution of each strategy is reported in Table~\ref{tab:improve} on the validation set. 

\vspace{\baselineskip}
\par\noindent\textbf{Inference time}
We tested the average run time on the Cityscapes validation set with a Nvidia Tesla V100 GPU. 
The Dilated FCN baseline and the Non-local model take 0.230s and 0.276s per image, respectively, while our proposed model uses 0.253s,
Thus, our proposed method is more efficient than Non-local~\cite{wang2018nonlocal} in execution time, FLOPs and also the number of parameters.

\vspace{\baselineskip}
\par\noindent\textbf{Effectiveness of different sampling rate $\varphi$ and group of predicted weights $G$ (Section 3.3 and 3.4 in main paper)}.
For our experiments on Cityscapes, where are network has a stride of 8, the sampling rates are set to  $\varphi = \{1, 6, 12, 24, 36\}$.
For experiments on COCO, where the network stride is 32, we use smaller sampling rates of $\varphi = \{1, 4, 8, 12\}$ in C5.
We keep the same sampling rate in C4 when DGMN modules are inserted into C4 as well.

Unless otherwise stated, all the experiments in the main paper and supplementary used $G=4$ groups as the default.
Each group of $C/G$ feature channels shares the same set of filter parameters~\cite{chollet2017xception}.

The effect of different sampling rates and groups of predicted filter weights are studied in Table~\ref{tab:segdilation}, for semantic segmentation on Cityscapes, and Table~\ref{tab:detdilation}, for object detection and instance segmentation on COCO.

\begin{table}[t]
	\tablestyle{2pt}{1.1}
	\centering
	\begin{tabular}{   l  |x{42}|x{52}|x{32}|x{42}}
		& OHEM & Multi-grid &MS &mIoU (\%)   \\
		\shline
		FCN w/ DGMN   &\xmark  &\xmark &\xmark & 79.2 \\
        FCN w/ DGMN & \cmark   & \xmark &\xmark  &79.7   \\
        FCN w/ DGMN &\cmark   &\cmark  & \xmark  &80.3   \\
        FCN w/ DGMN &\cmark  &\cmark & \cmark   &\bf 81.1   \\
	\end{tabular}
	\caption{\small
	Ablation studies of different training and inference strategies.
Our method (DGMN w/ DA+DW+US) is evaluated under single scale mIoU with ResNet-101 backbone on Cityscapes validation set.
}\label{tab:improve}
\end{table}

\begin{table*}[t]
\tablestyle{2pt}{1.2}
	\centering
	\begin{tabular}{   l  |x{32}|x{32}|x{32}|x{32}}
     &DA &DW& DS & mIoU (\%)   \\
    \shline
    Dilated FCN    &\xmark &\xmark &\xmark & 75.0 \\
    \hline
    + DGMN ($\varphi =\{1\}$)  & \cmark &\xmark &\xmark  &  76.5 \\
    + DGMN ($\varphi =\{1\}$)  & \cmark & \cmark  &\xmark &  79.1 \\
    + DGMN ($\varphi = \{1, 1, 1, 1\}$)  & \cmark & \cmark &\cmark & 79.2 \\
    + DGMN ($\varphi = \{1, 6, 12\}$)  & \cmark & \cmark & \cmark  &  79.7 \\
    + DGMN ($\varphi = \{1, 6, 12, 24, 36\}$)  & \cmark & \cmark &\cmark  &\bf 80.4  \\
    \end{tabular}
\caption{\small 
Quantitative analysis on different sampling rates of our dynamic sampling strategy in the proposed DGMN model on the Cityscapes validation set. 
We report the mean IoU and use a ResNet-101 as backbone. 
All methods are evaluated using a single scale.}
\label{tab:segdilation}
\end{table*}


\begin{table*}[h!]
\tablestyle{2pt}{1.2}
	\centering
    \begin{tabular}{   l  |x{32}|x{32}|x{32}|x{32}|x{32}}
     &DA &DW & DS & $\mathrm{AP^{box}}$ & $\mathrm{AP^{mask}}$  \\
    \shline
    Mask R-CNN baseline   &\xmark &\xmark &\xmark  & 37.8 & 34.4\\
    \hline
    + DGMN ($\varphi = \{1, 4, 8, 12\}$, $G = 0$)  & \cmark & \xmark &\xmark &39.4 &35.6 \\
    + DGMN ($\varphi = \{1, 4, 8, 12\}$, $G = 0$)  & \cmark & \xmark &\cmark &39.9 &35.9 \\
    + DGMN ($\varphi = \{1, 4, 8\}$, $G = 2$)  & \cmark & \cmark  & \cmark &39.5 &35.6 \\
    + DGMN ($\varphi = \{1, 4, 8\}$, $G = 4$)  & \cmark & \cmark  & \cmark &39.8 &35.9  \\
    + DGMN ($\varphi = \{1, 4, 8, 12\}$, $G = 4$)  & \cmark & \cmark & \cmark  &\bf 40.2  &\bf 36.0 \\
    \end{tabular}
\caption{\small
Quantitative analysis on different numbers of filter groups ($G$) and sampling rates ($\varphi$) for the proposed DGMN model on the COCO 2017 validation set. 
All methods are based on the Mask R-CNN detection pipeline with a ResNet-50 backbone, and evaluated on the COCO validation set.
Modules are inserted after all the $3 \times 3$ convolution layers of C5 (\emph{res5}) of ResNet-50.}
\label{tab:detdilation}
\end{table*}

\vspace{\baselineskip}
\par\noindent\textbf{Effectiveness of feature learning with DGMN on stronger backbones.}
Table 4 of the main paper showed that our proposed DGMN module still provided substantial benefits on the more powerful backbones such as ResNet-101 and ResNeXt 101 on the COCO test set.
Table~\ref{tab:strongbackbone} shows this for the COCO validation set as well.
By inserting DGMN at the convolutional stage C5 of ResNet-101, DGMN (C5) outperforms the Mask R-CNN baseline with 1.6 points on the  AP$^{\mathrm{box}}$ metricand by 1.2 points on the AP$^{\mathrm{mask}}$ metric. 
On ResNeXt-101, DGMN (C5) also improves by 1.5 and 0.9 points on the AP$^{\mathrm{box}}$ and the AP$^{\mathrm{mask}}$, respectively.

\begin{table*}[h!]
\tablestyle{2pt}{1.2}
	\centering
	\begin{tabular}{   l  |x{52}|x{32}x{32}x{32}|x{32}x{32}x{32}}
Model & Backbone & $\mathrm{AP^{box}}$ & $\mathrm{AP^{box}_{50}}$ & $\mathrm{AP^{box}_{75}}$ &$\mathrm{AP^{mask}}$ &$\mathrm{AP^{mask}_{50}}$ &$\mathrm{AP^{mask}_{75}}$  \\
\shline
Mask R-CNN baseline & \multirow{2}{*}{ResNet-101} &40.1 &61.7 &44.0 &36.2  &58.1 &38.3 \\
+ DGMN (C5) &  &\bf 41.7 &\bf 63.8 &\bf 45.7 &\bf 37.4  &\bf 60.4 &\bf 39.8 \\
\hline 
Mask R-CNN baseline & \multirow{2}{*}{ResNeXt-101}   &42.2 &63.9 &46.1 &37.8  &60.5 &40.2 \\
+ DGMN (C5)  &  &\bf 43.7 &\bf 65.9 &\bf 47.8 &\bf 38.7  &\bf 62.1 &\bf 41.3 \\
\end{tabular}
\caption{\small
Quantitative results via applying the proposed DGMN module into different strong backbone networks for object detection and instance segmentation on the COCO 2017 validation set.}
\label{tab:strongbackbone}
\end{table*} 

\subsection{State-of-the-art comparison on COCO}
\label{sec:coco_sota}
Table~\ref{tab:coco_sota} shows comparisons to the state-of-the-art on the COCO \texttt{test-dev} set.
When testing, we process a single scale using a single model.
We do not perform any other complementary performance-boosting ``tricks''.
Our DGMN approach outperforms one-stage detectors including the most recent CornerNet~\cite{law2018cornernet} by 2.1 points on box Average Precision (AP). 
DGMN also shows superior performance compared to two-stage detectors including Mask R-CNN~\cite{maskrcnn} and Libra R-CNN \cite{pang2019libra} using the same ResNeXt-101-FPN backbone.

\begin{table*}[!t]
	\tablestyle{2pt}{1.1}
	\centering
	\begin{tabular}{   l  |c|x{22}x{22}x{32}|x{32}x{32}x{32}}
		& Backbone
		& $\mathrm{AP^{box}}$ & $\mathrm{AP^{box}_{50}}$ & $\mathrm{AP^{box}_{75}}$ &$\mathrm{AP^{mask}}$ &$\mathrm{AP^{mask}_{50}}$ &$\mathrm{AP^{mask}_{75}}$  \\ [.1em]
		\shline
		\emph{One-stage detectors} & & & & & & & \\
		~YOLOv3 \cite{redmon2018yolov3} & Darknet-53 & 33.0 &57.9 &34.4 & -& - & - \\
		~SSD513 \cite{liu2016ssd} & ResNet-101-SSD & 31.2 & 50.4 & 33.3 & - & - & - \\
		~DSSD513 \cite{fu2017dssd} & ResNet-101-DSSD & 33.2 & 53.3 & 35.2 & - & - & - \\
		~{RetinaNet} \cite{lin2017focal} & ResNeXt-101-FPN &40.8 &61.1 &44.1 & & & \\
		~CornerNet \cite{law2018cornernet} & Hourglass-104 &42.2 &57.8 &45.2 & & &    \\
		\hline
		\emph{Two-stage detectors} & & & & & & & \\
		~Faster R-CNN+++ \cite{resnet} & ResNet-101-C4 & 34.9 & 55.7 & 37.4 &- &- & -\\
		~Faster R-CNN w FPN \cite{fpn} & ResNet-101-FPN & 36.2 & 59.1 & 39.0 & - & - & -\\
		~R-FCN \cite{dai2016r} & ResNet-101 & 29.9 & 51.9  & - & - & - & - \\
		~Mask R-CNN \cite{maskrcnn} & ResNet-101-FPN &40.2 &61.9 &44.0 &36.2  &58.6 &38.4 \\
	    ~Mask R-CNN \cite{maskrcnn}  & ResNeXt-101-FPN &42.6 &64.9 &46.6 &38.3  &61.6 &40.8 \\
	    ~Libra R-CNN \cite{pang2019libra}& ResNetX-101-FPN  & 43.0 &64.0 &47.0 & - & - & - \\
		\hline 
		~\textbf{DGMN} (ours) & ResNeXt-101-FPN&\bf 44.3   &\bf 66.8 &\bf 48.4  &\bf 39.5 &\bf 63.3 &\bf 42.1 \\
	\end{tabular}
	\caption{
		Object detection and instance segmentation performance using a \textit{single-model} on the COCO \texttt{test-dev} set. 
		We use \textit{single scale} testing.
	}\label{tab:coco_sota}
\end{table*}

\input{figures/cs_results}
\input{figures/coco_results}

\clearpage

%% file: figures/cs_results.tex
\begin{figure*}[h]
	\centering
	\includegraphics[width=1.0\linewidth]{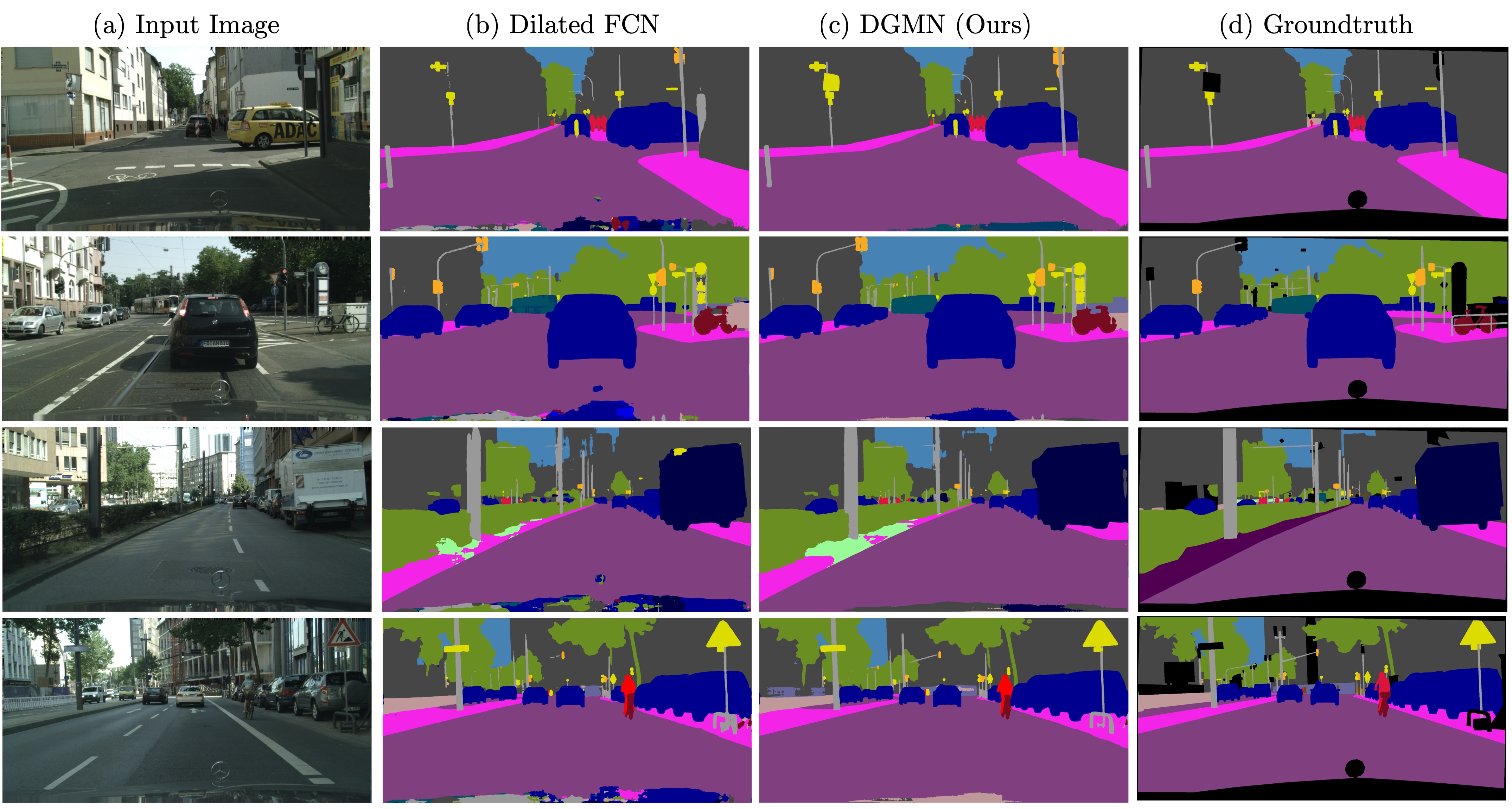}
	\caption{Qualitative results of the Dilated FCN baseline~\cite{yu2015multi} and our proposed DGMN model on the Cityscapes dataset.}
	\label{fig:cs_results}
\end{figure*}

%% file: figures/coco_results.tex
\begin{figure*}[htb]
    \vspace{-\baselineskip}
	\def \imheight {3.6cm}
	\centering
	\rotatebox{90}{\textcolor{white}{------}Mask R-CNN} \hspace{0.05cm}
	\includegraphics[height=\imheight, align=b]{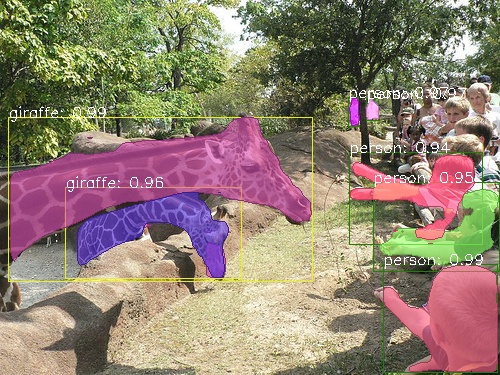}
	\includegraphics[height=\imheight, align=b]{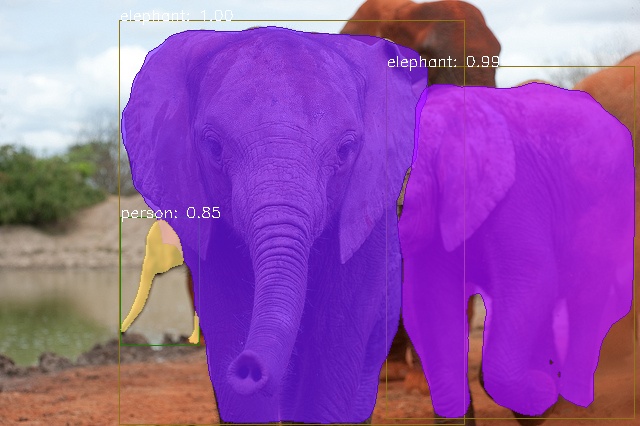}
	\includegraphics[height=\imheight, align=b]{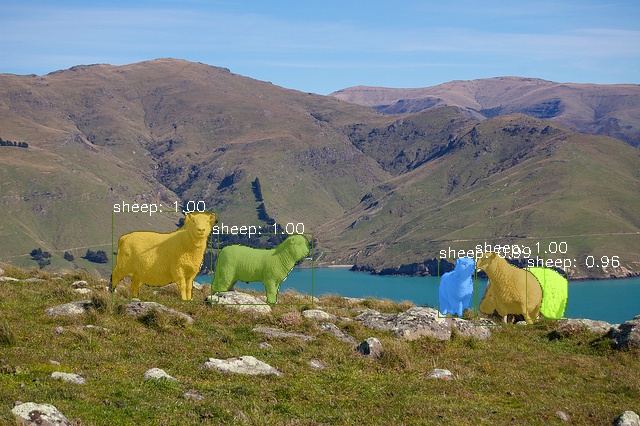}
	\\
	\rotatebox{90}{\textcolor{white}{------}DGMN (ours)} \hspace{0.05cm}
	\includegraphics[height=\imheight, align=b]{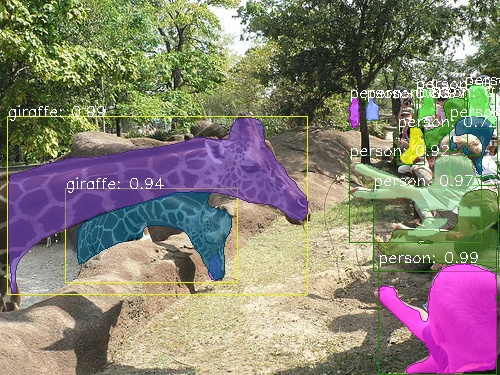}
	\includegraphics[height=\imheight, align=b]{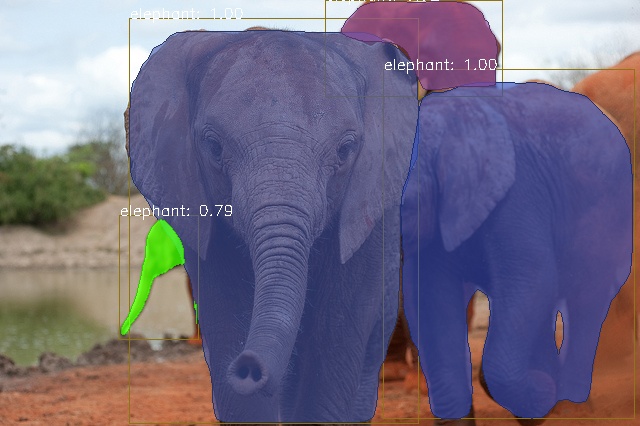}
	\includegraphics[height=\imheight, align=b]{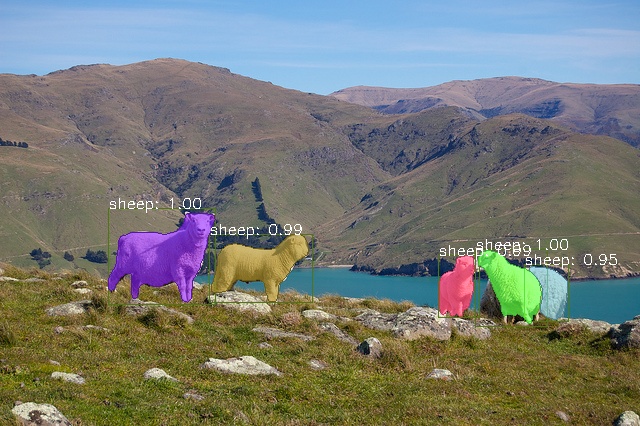}
	\\
	
	\rotatebox{90}{\textcolor{white}{------}Mask R-CNN} \hspace{0.05cm}
	\includegraphics[height=\imheight, align=b]{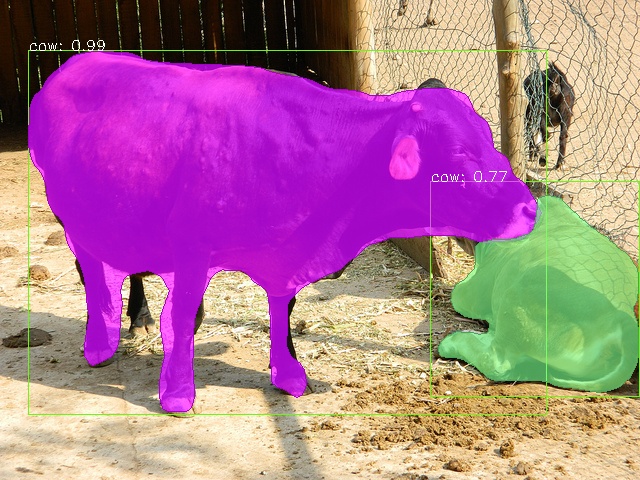}
	\includegraphics[height=\imheight, align=b]{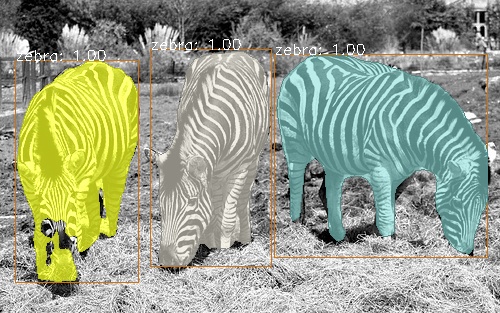}
	\includegraphics[height=\imheight, align=b]{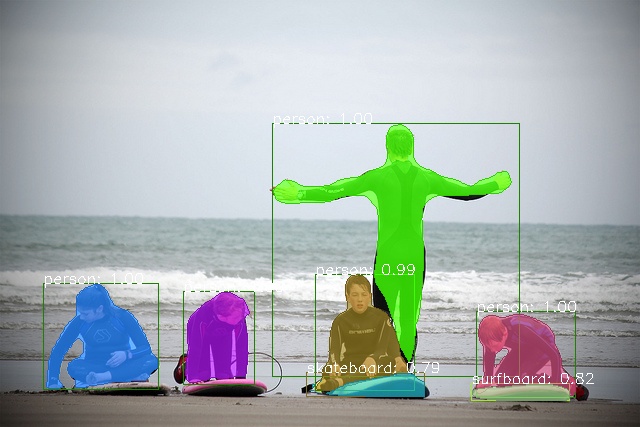}
	\\
	\rotatebox{90}{\textcolor{white}{------}DGMN (ours)} \hspace{0.05cm}
	\includegraphics[height=\imheight, align=b]{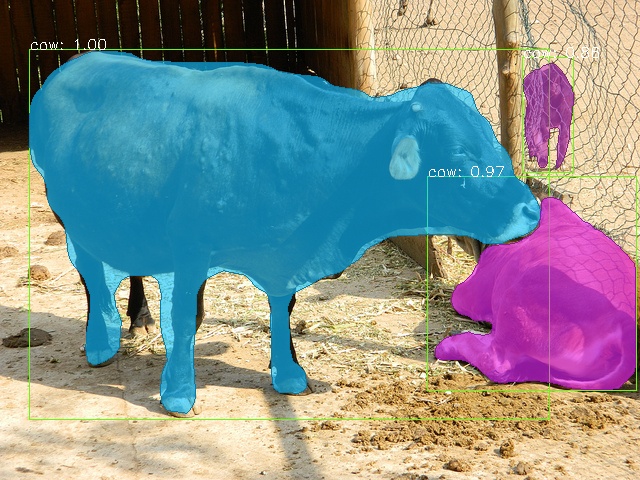}
	\includegraphics[height=\imheight, align=b]{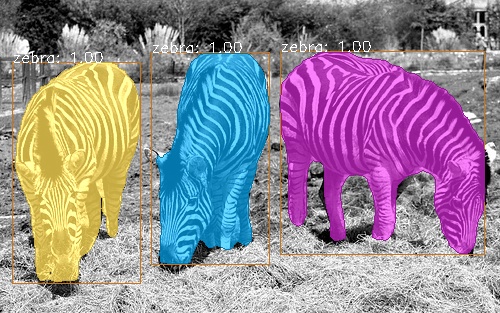}
	\includegraphics[height=\imheight, align=b]{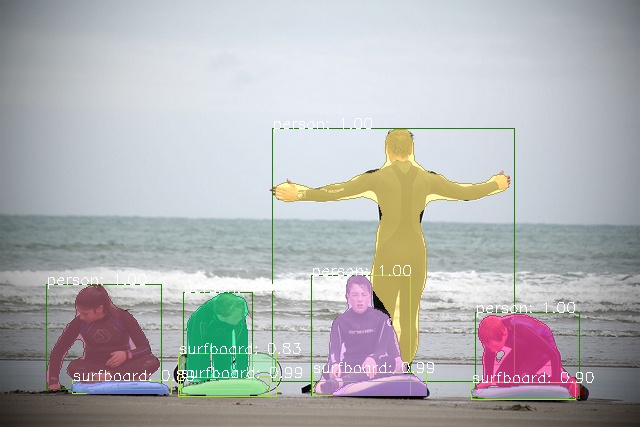}
	\\
	
	\rotatebox{90}{\textcolor{white}{------}Mask R-CNN} \hspace{0.05cm}
	\includegraphics[height=\imheight, align=b]{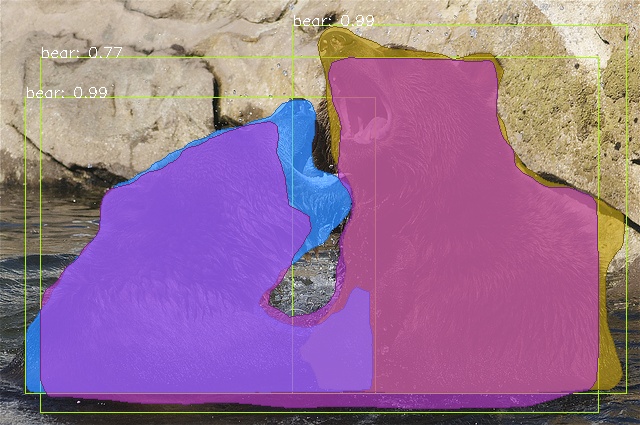}
	\includegraphics[height=\imheight, align=b]{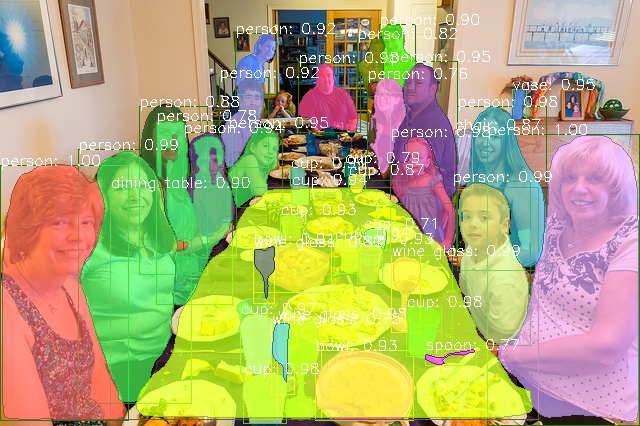}
	\includegraphics[height=\imheight, align=b]{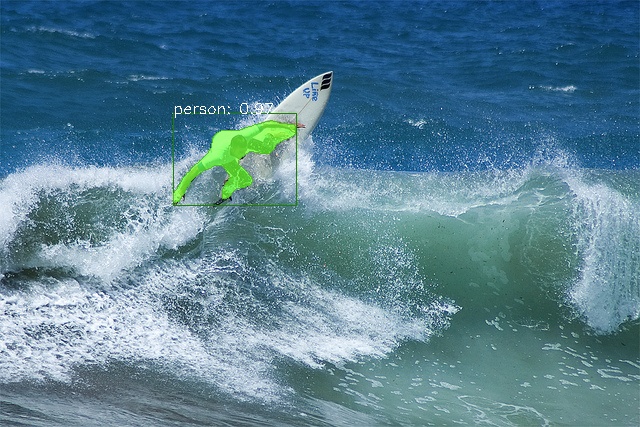}
	\\
	\rotatebox{90}{\textcolor{white}{------}DGMN (ours)} \hspace{0.05cm}
	\includegraphics[height=\imheight, align=b]{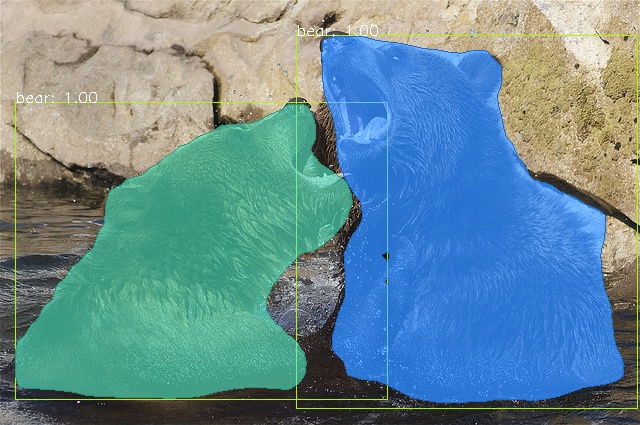}
	\includegraphics[height=\imheight, align=b]{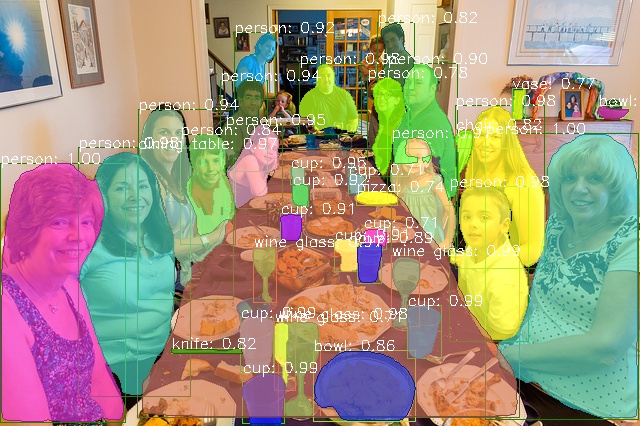}
	\includegraphics[height=\imheight, align=b]{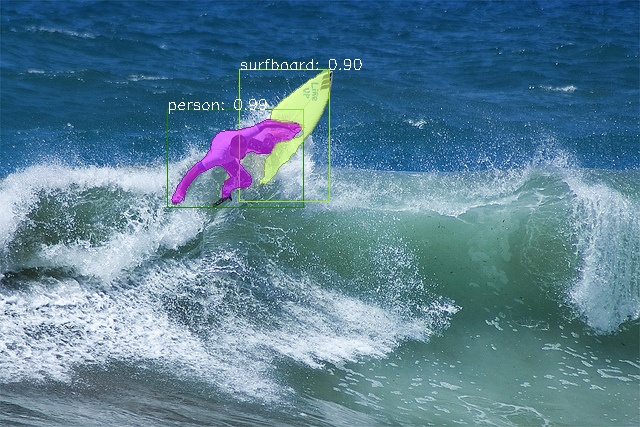}
	\\
	
	\caption{Qualitative examples of the instance segmentation task on the COCO validation dataset. 
	The odd rows are the results from the Mask R-CNN baseline~\cite{massa2018mrcnn,maskrcnn}. 
	The even rows are the results from our DGMN approach.
	Note how our approach often produces better segmentations and fewer false-positive and false-negative detections.
	}
	\label{fig:coco}
\end{figure*}

\begin{figure*}[htb]
    \vspace{-\baselineskip}
	\def \imheight {3.6cm}
	\centering
	
	\rotatebox{90}{\textcolor{white}{------}Mask R-CNN} \hspace{0.05cm}
	\includegraphics[height=\imheight, align=b]{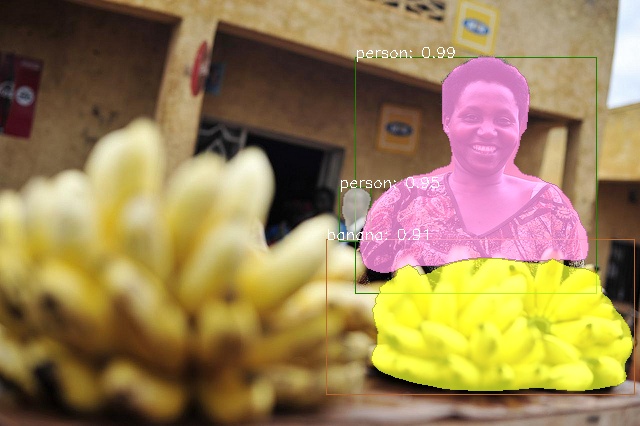}
	\includegraphics[height=\imheight, align=b]{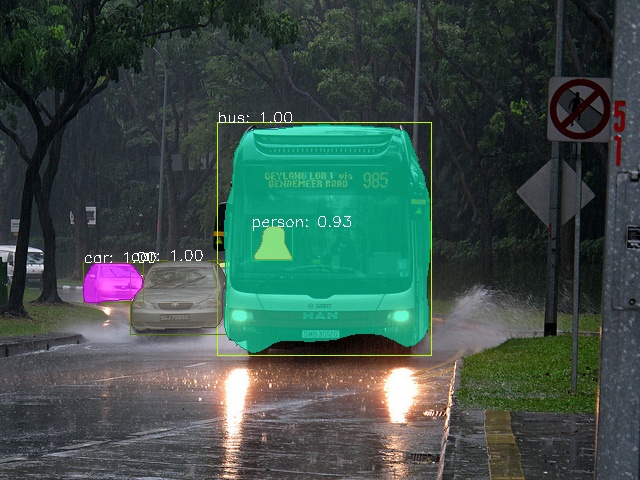}
	\includegraphics[height=\imheight, align=b]{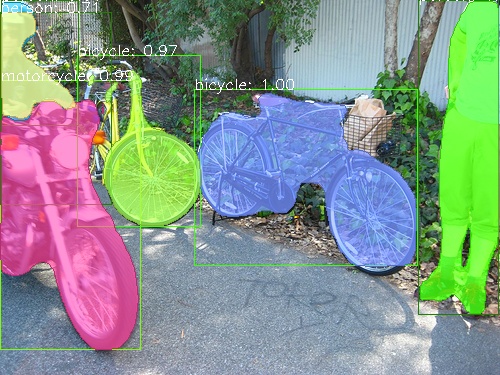}
	\\
	\rotatebox{90}{\textcolor{white}{------}DGMN (ours)} \hspace{0.05cm}
	\includegraphics[height=\imheight, align=b]{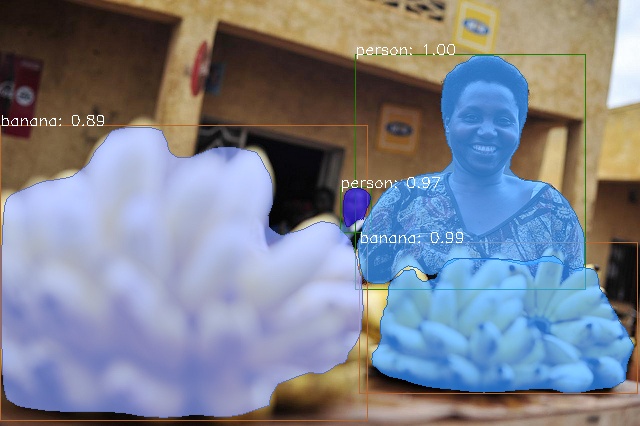}
	\includegraphics[height=\imheight, align=b]{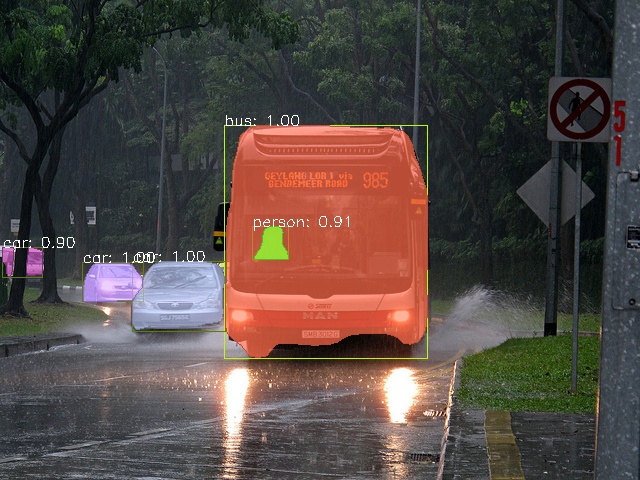}
	\includegraphics[height=\imheight, align=b]{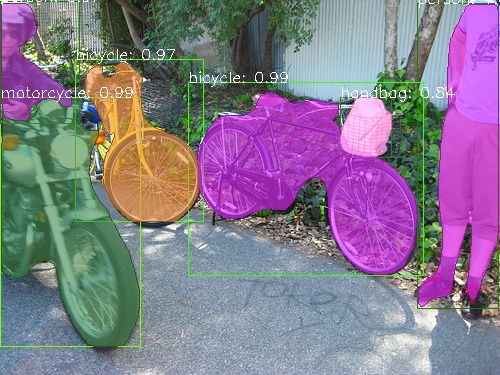}
	\\
	
	\rotatebox{90}{\textcolor{white}{------}Mask R-CNN} \hspace{0.05cm}
	\includegraphics[height=\imheight, align=b]{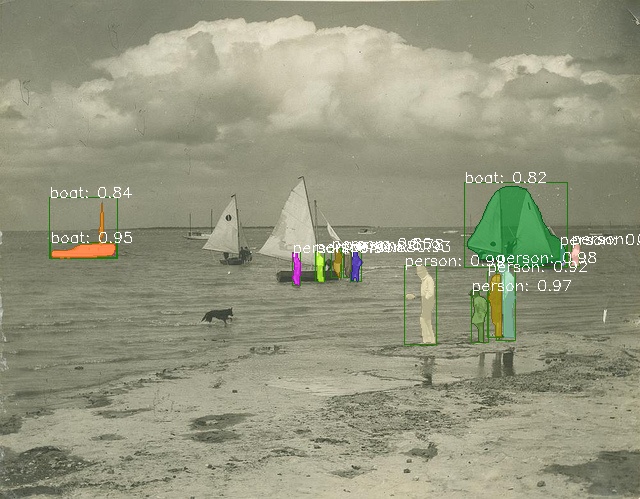}
	\includegraphics[height=\imheight, align=b]{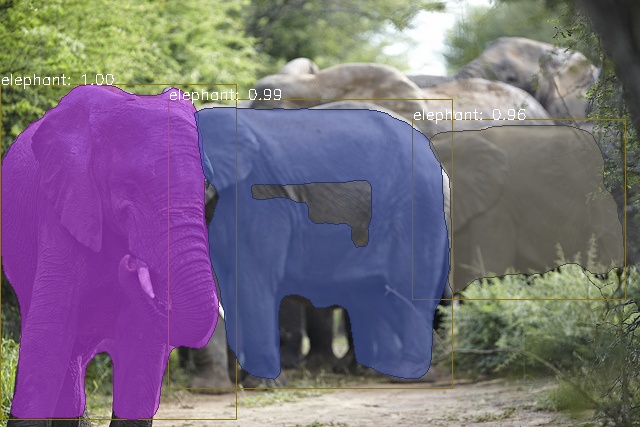}
	\includegraphics[height=\imheight, align=b]{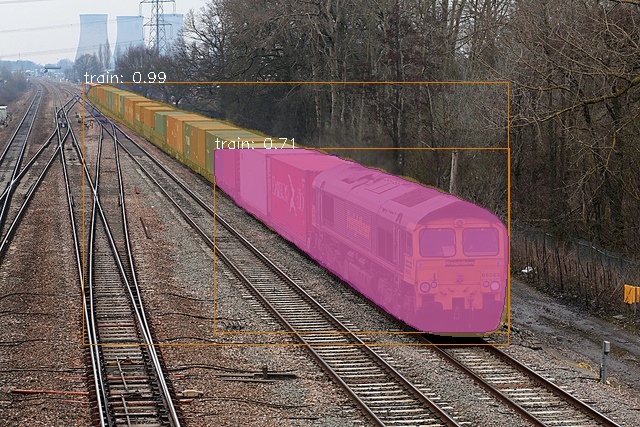}
	\\
	\rotatebox{90}{\textcolor{white}{------}DGMN (ours)} \hspace{0.05cm}
	\includegraphics[height=\imheight, align=b]{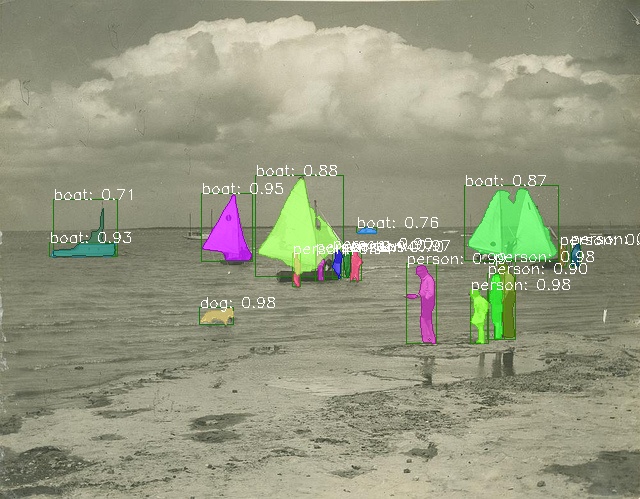}
	\includegraphics[height=\imheight, align=b]{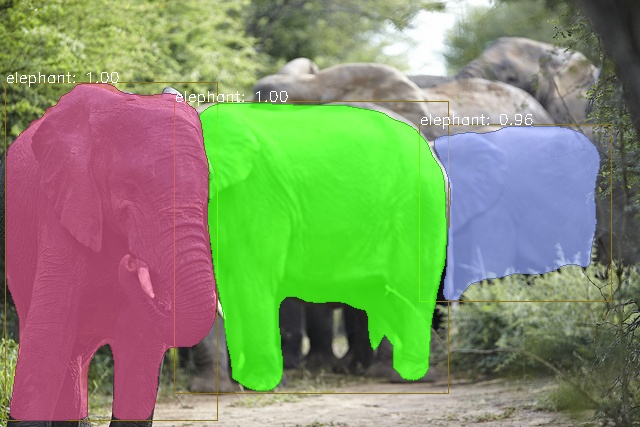}
	\includegraphics[height=\imheight, align=b]{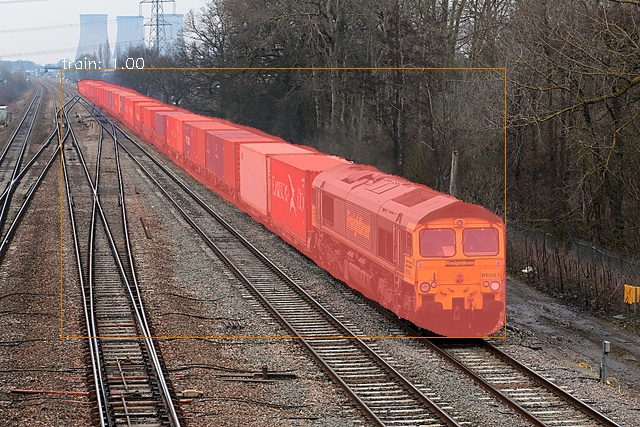}
	\\
	
	\rotatebox{90}{\textcolor{white}{------}Mask R-CNN} \hspace{0.05cm}
	\includegraphics[height=\imheight, align=b]{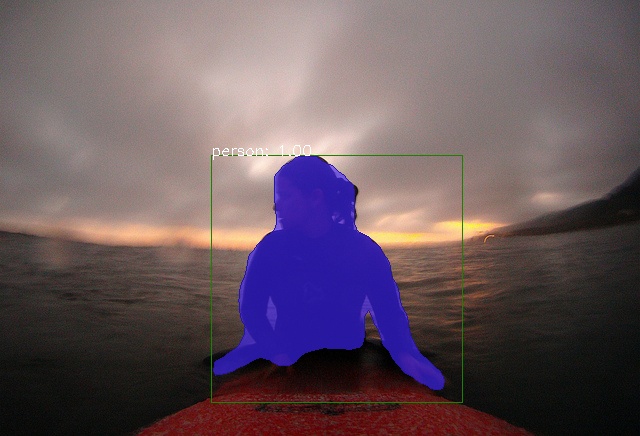}
	\includegraphics[height=\imheight, align=b]{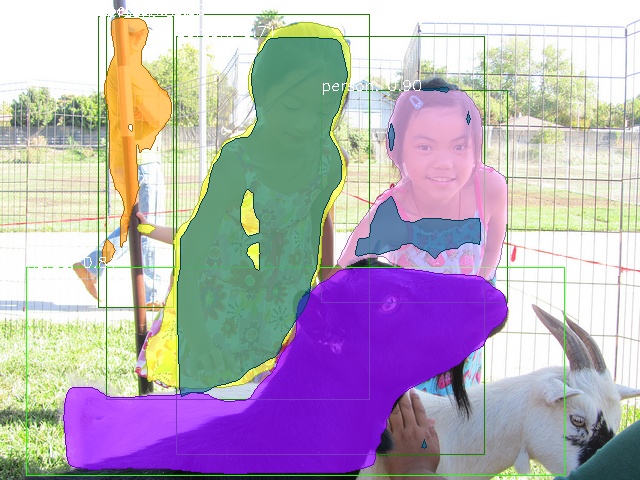}
	\includegraphics[height=\imheight, align=b]{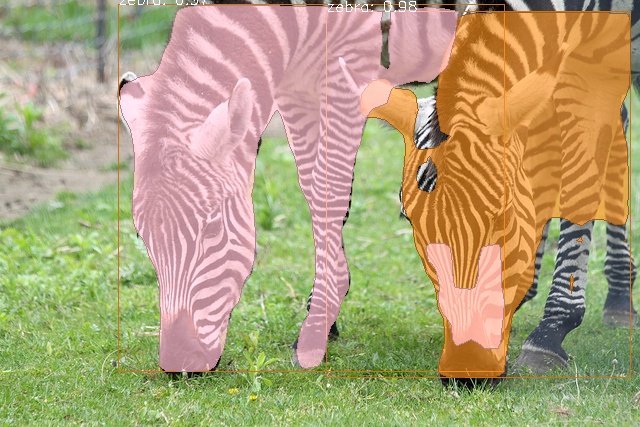}
	\\
	\rotatebox{90}{\textcolor{white}{------}DGMN (ours)} \hspace{0.05cm}
	\includegraphics[height=\imheight, align=b]{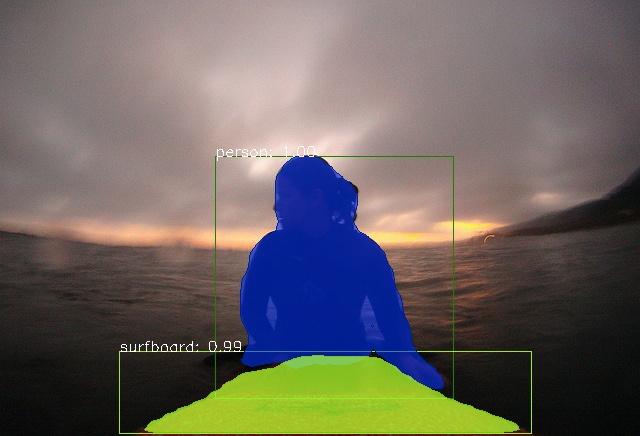}
	\includegraphics[height=\imheight, align=b]{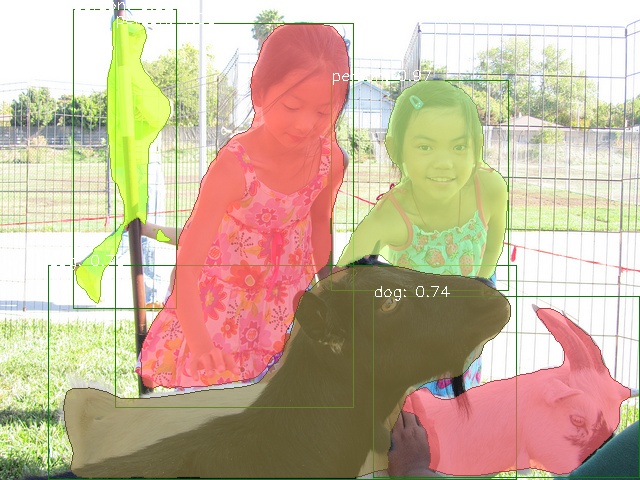}
	\includegraphics[height=\imheight, align=b]{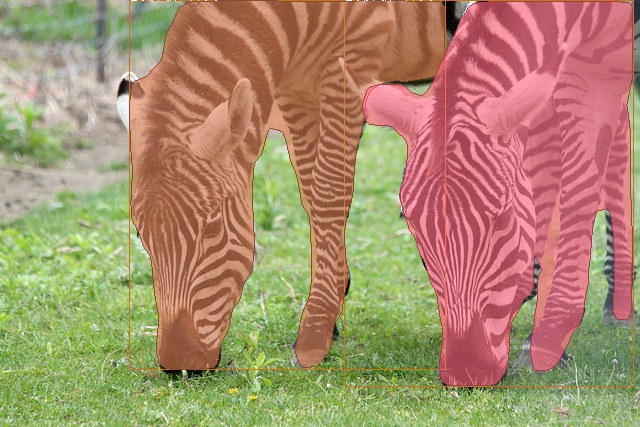}

	\caption{More qualitative examples of the instance segmentation task on the COCO validation dataset. 
	The odd rows are the results from the Mask R-CNN baseline~\cite{massa2018mrcnn,maskrcnn}. 
	The even rows are the detection results from our DGMN approach.
	Note how our approach often produces better segmentations and fewer false-positive and false-negative detections.
	}
	\label{fig:coco2}
\end{figure*}